\documentclass[11pt,a4paper]{article}
\usepackage{times,latexsym}
\usepackage{url}
\usepackage[T1]{fontenc}

\usepackage[acceptedWithA]{tacl2021v1}
\usepackage{xspace,mfirstuc,tabulary}

\newif\iftaclinstructions
\taclinstructionsfalse % AUTHORS: do NOT set this to true
\iftaclinstructions
\renewcommand{\confidential}{}
\renewcommand{\anonsubtext}{(No author info supplied here, for consistency with
TACL-submission anonymization requirements)}
\newcommand{\instr}
\fi

\iftaclpubformat % this "if" is set by the choice of options

\else

\fi

\usepackage[utf8]{inputenc} % allow utf-8 input
\usepackage[T1]{fontenc}    % use 8-bit T1 fonts
\usepackage{hyperref}       % hyperlinks
\usepackage{url}            % simple URL typesetting
\usepackage{booktabs}       % professional-quality tables
\usepackage{amsfonts}       % blackboard math symbols
\usepackage{nicefrac}       % compact symbols for 1/2, etc.
\usepackage{microtype}      % microtypography
\usepackage[table]{xcolor}         % colors

\usepackage{multirow}
\usepackage{enumerate}
\usepackage{nicematrix}
\usepackage{arydshln}
\usepackage{amsmath} % For better math support
\usepackage{bm} % For bold math symbols
\usepackage{caption}
\usepackage{ulem}           % for \normalem
\usepackage[dvipsnames]{xcolor}
\usepackage{float}

\renewcommand{\emph}[1]{{\it #1}}

\newcommand{\revision}[1]{{\color{black}{#1}}}

\newcommand{\diff}[1]{{\tiny\color{blue}{#1}}}

\title{Triospect: A Three-Dimensional Framework for Robust Statistical AI-Generated Text Detection Against Diverse Attacks}

\author{
    Guangsheng Bao\textsuperscript{\rm 1,3,\footnotemark[1]},
    Lihua Rong\textsuperscript{\rm 2,\footnotemark[1]},
    Yanbin Zhao\textsuperscript{\rm 4},
    Xiao Yu\textsuperscript{\rm 5},
    Qiji Zhou\textsuperscript{\rm 3}, 
    and Yue Zhang\textsuperscript{\rm 3,\footnotemark[2]}
    \\
    \\
    \textsuperscript{1} Zhejiang University \hspace{15pt} 
    \textsuperscript{2} Zhejiang University of Technology  \hspace{15pt}      \textsuperscript{3} Westlake University \\
    \textsuperscript{4} Shanghai Polytechnic University \hspace{15pt}    
    \textsuperscript{5} University of Science and Technology of China \\
    \texttt{baoguangsheng@westlake.edu.cn}\footnotemark[1] \hspace{15pt} 
    \texttt{ronglihua1981@zjut.edu.cn}\footnotemark[1] \\
    \texttt{zhangyue@westlake.edu.cn}\footnotemark[2] \\
}
\date{}

\begin{document}
\maketitle

\renewcommand{\thefootnote}{\fnsymbol{footnote}}
\footnotetext[1]{Equal contribution. \footnotemark[2]Corresponding author.}
\renewcommand{\thefootnote}{\arabic{footnote}}

\begin{abstract}
Existing AI-generated text detectors are vulnerable to attacks that manipulate textual characteristics. In this study, we propose a novel \emph{Triospect Detection Framework} by using additional perspectives of content (core ideas) and expression (stylistic elements) within a given text. Experiments on two benchmarks involving 17 attacks, 12 domains, and 17 source models demonstrate that Triospect is robust against these attacks. It improves the strong baseline by a significant margin of 22.3\% (AUROC) and 13\% (TPR01) on the Humanize-16K after-attack subset, and by 9.1\% (AUROC) and 22\% (TPR01) on the adversarial RAID. This framework marks a pioneering effort in \revision{statistical} methods to enhance detection reliability against attacks. We release our data and code at \url{https://github.com/baoguangsheng/triospect}.
\end{abstract}

\section{Introduction}
% 1) LLM lead to risks -> detectors -> humanizing tools to by-pass -> create new threats. (Background)
Large language models (LLMs) have been widely used in news, academic, story and advertising writing \cite{christian2023cnet, m2022exploring, yuan2022wordcraft, chen2023large}, leading to unprecedented societal risks including spreading misinformation \cite{ahmed2021detecting,bagdasaryan2022spinning,chen2023combating}, eroding academic integrity \cite{perkins2023academic,lee2023language,kumar2024academic}, and blurring accountability in digital communication \cite{kaur2022trustworthy,sun2024trustllm}. These challenges call for reliable AI-generated text detection tools, requiring the research community to develop effective detectors. However, at the same time, attacking techniques are also developed \cite{krishna2024paraphrasing,zhou2024humanizing,ayub2024art}, and commercial AI humanizing tools (16 listed in Appendix \ref{app:external_tools}) can bypass state-of-the-art detectors, posing urgent threats to safe usage of AI technology.

Existing state-of-the-art detectors are vulnerable when facing attacks.  Supervised detectors \cite{solaiman2019release,fagni2021tweepfake,yan2023detection,li2024mage,verma2024ghostbuster} may fail in unfamiliar synthesized styles \cite{bakhtin2019real,uchendu2020authorship,pu2023deepfake}. Zero-shot detectors \cite{gehrmann2019gltr,su2023detectllm, mitchell2023detectgpt, bao2024fast,xu2024detecting,hans2024spotting} may be affected by surface-level textual changes \cite{krishna2024paraphrasing,dugan2024raid,wu2024detectrl,chen2025imitate}. The watermarks \cite{kirchenbauer2023watermark,zhao2023protecting,christ2024undetectable, zhao2024permute,zhao2024sok} can be removed using new words and syntax \cite{krishna2024paraphrasing,sadasivan2023can}. 
These vulnerabilities underscore the necessity of advancing robust detection mechanisms that can withstand a wide range of attacks.

\begin{figure*}[t]
    % \vspace{-0.2in}
    \centering
    \includegraphics[trim={20pt 0pt 10pt 0pt},clip,width=1.0\linewidth]{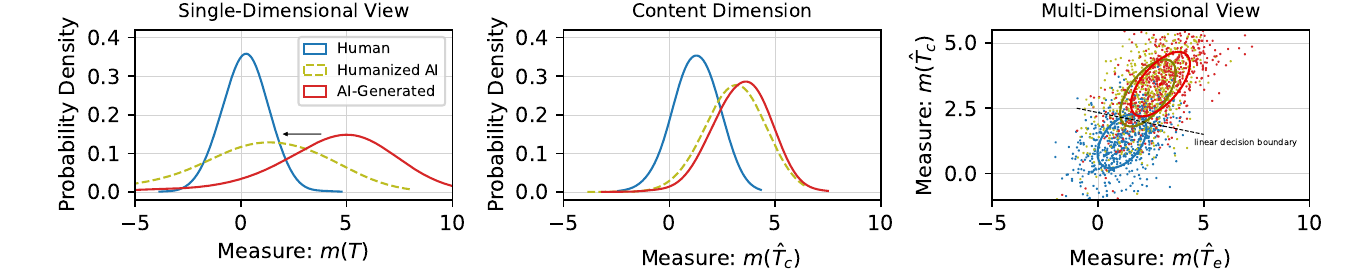}   
    \caption{\textbf{Left:} Humanizing attacks cause failure of existing detectors in \emph{single-dimensional view}. \textbf{Middle:} The failure can be mitigated by more stable content dimension. \textbf{Right:} In \emph{multi-dimensional view}, we combine the dimensions to achieve robust detection. (Humanize-16K using Fast-Detect as the metric)}
    \label{fig:single_vs_2d}
    % \vspace{-0.2in}
\end{figure*}

\revision{Such attacks generally modify textual expressions (surface forms) while largely preserving underlying contents (semantics)~\citep{jia2017adversarial, alzantot2018generating, ribeiro2018semantically, jin2020bert}}. By \emph{content}, we refer to the information conveyed in the text. It is about `what' is being communicated. By \emph{expression}, we refer to the manner in which the content is conveyed. It is about `how' the content is communicated.
That is to say\revision{, under} these attacks, the content of a text remains relatively stable. 
\revision{Thus,} if a detector can measure the content directly, it will be resilient against the attacks\revision{, as adversarial vulnerability often stems from reliance on non-robust surface features rather than semantically meaningful representations~\citep{madry2017towards, ilyas2019adversarial}.} 
To separate the content from the expression of a text, we leverage the strong understanding ability of LLMs to regularize its expression, leading to a resultant token sequence primarily determined by its content. This content-representing token sequence has unique features for identifying its source, which we find can be captured by existing detection metrics. 

Based on this idea, we propose \textbf{Triospect}, a three-dimensional \revision{statistical detection framework that jointly analyzes the original text, its content, and its expression, enabling complementary signals to be examined across these three distinct dimensions.}
\revision{Specifically, given a text $T$,} we perform a textual transformation to convert it into an \revision{approximate content-preserving text $\hat{T}_c$ and expression-preserving text $\hat{T}_e$}, approximately decoupling the two aspects. 
We measure the texts using an existing detection metric $m(\cdot)$ (e.g., Fast-Detect), obtaining the \emph{original text measure} $m(T)$, the \emph{content measure} \revision{$m(\hat{T}_c)$}, and the \emph{expression measure} \revision{$m(\hat{T}_e)$}. 
As Figure \ref{fig:single_vs_2d} illustrates, humanizing tools shift the distribution of AI-generated texts towards human-written texts in the measure $m(T)$, causing overlap between `{\it Human}' and `{\it Humanized AI}'. However, this confusion can be mitigated by the \revision{content measure}, where the distribution remains relatively stable under attacks. Additionally, the \revision{expression measure} can help when an attack occurs at the word level and does not change the grammatical patterns. Consequently, by combining these measures in the multi-dimensional view, we achieve a more robust detection framework.

% 5) Specifically, content-preserved texts -> expression-preserved texts -> content measure and expression measure 
% Together with the \emph{original text measure} $m(T)$, we create a three-dimensional view $[m(T), m(T_c), m(T_e)]$. We model the vector in a multivariate Gaussian mixture distribution and predict the probability of a text being AI-generated by contrasting its vector probability density in the machine distribution with the sum of the machine and human probability densities.

We construct a \emph{Humanize-16K} benchmark for the systematic evaluation of detectors dealing with humanizing attacks. We collect human-written texts from 4 data sources and generate corresponding AI texts based on the same contexts (titles or prompts) using 6 LLMs. Then, we apply 6 humanizing attacks to the AI-generated texts, where the attacks include human editing, commercial AI tools, and LLM simulated attacks. Consequently, we obtain 16K samples, half for development and half for testing.

% 6) Results: resist to humanizing attack -> resist to adversarial attack -> obtain improved results.
We evaluate detectors on our Humanize-16K and the existing adversarial RAID benchmarks, where they cover a total of 12 domains, 17 LLMs, and 17 attacks. The experimental results show that Triospect is resistant to humanizing and adversarial attacks, leading to an improvement of 22.3\% in AUROC and 13\% in TPR01 on the Humanize-16K after-attack subset, as well as an improvement of 9.1\% in AUROC and 22\% in TPR01 on RAID, compared to the strong baseline detector. Analysis shows that Triospect is also robust to source models, decoding strategies, text lengths, and languages.

% 7) Conclusion: content measure significantly improves the ability of existing detectors in real world. 
In summary, perspectives from content and expression provide beneficial complements to original texts, leading to more reliable detection under humanizing and adversarial attacks. To our knowledge, this is \emph{the first \revision{statistical} method to tackle detection problems against a wide range of attacks, and it achieves new state-of-the-art results among zero-shot detectors.}

\section{Related Work}

\paragraph{AI-Generated Text Detection}

Existing detectors consist of three types of technology. The first is supervised classifiers \cite{solaiman2019release,ippolito2020automatic,fagni2021tweepfake,hu2023radar,yan2023detection,li2024mage,verma2024ghostbuster,yu2024dpic,chen2025imitate}, which train a binary classifier based on a large collection of AI-generated and human-written text. The second is zero-shot classifiers, including white-box methods \cite{gehrmann2019gltr,su2023detectllm,bao2024fast,xu2024detecting,hans2024spotting} and black-box methods \cite{mitchell2023detectgpt,yang2023dna,bhattacharjee2024fighting,bao2025glimpse,chen2025imitate}. These technologies usually use pre-trained language models to extract detection metrics. The third is text watermarking technology \cite{kirchenbauer2023watermark,zhao2023protecting,christ2024undetectable, zhao2024permute,zhao2024sok}, which identifies AI-generated text by embedding easy-to-detect markers or patterns.
While these methods are successful at identifying entirely AI-generated texts, they lack resilience against diverse attack strategies \cite{gao2018black,dyrmishi2023humans,krishna2024paraphrasing,he2024mgtbench,dugan2024raid,wu2024detectrl,wang2024stumbling,zhou2024humanizing}. As a result, various commercial AI tools offer features to ``humanize'' content, enabling them to evade current detection systems. To tackle this issue, we introduce the Triospect detection framework as a promising solution to counteract such vulnerabilities. 

There is limited research focused on the humanizing attack, with most efforts relying on supervised learning. For instance, ImBD \cite{chen2025imitate} fine-tunes a LLM on AI-rewritten and refined texts to learn machine-generated styles, using a style-CPC (conditional probability curvature) metric for detection. Similarly, DAMAGE \cite{masrour2025damage} develops a binary classifier trained on humanized texts produced by various humanizers. In contrast, Triospect does not require training and can be integrated with existing detectors to enhance their performance, including the supervised ImBD.

The existing rewrite-detect frameworks \cite{mitchell2023detectgpt,liu2024does,maoraidar2024} \revision{and recent repair-based DNA-DetectLLM \cite{zhudna2025}} may appear superficially similar to our Triospect framework since both employ LLMs to \revision{transform} original texts. However, they differ fundamentally in their hypotheses and principles. The Triospect framework is driven by the idea of separating the content and expression of a text. We propose that content and expression are two crucial aspects of a text, offering stable and clear signals to identify AI sources. By decoupling these elements, we can evaluate the content and expression dimensions independently. In contrast, the existing rewrite-detect frameworks generally assume that AI-generated texts, when rewritten, exhibit greater similarity. They measure the similarity between rewritten versions, for example, using 7 versions by Raidar \cite{maoraidar2024}, as a means to indicate AI generation.

\paragraph{Decoupling Content and Expression}
The idea of separating content and expression is related to existing studies on the disentanglement of semantics and syntax. These studies mainly focus on the disentanglement at the sentence level and discuss it in different contexts, such as recognition science \cite{caucheteux2021disentangling,moro2001syntax}, sentence representation \cite{chen2019multi}, sentence comprehension \cite{dapretto1999form}, and sentence generation \cite{bao2019generating}. They generally represent semantics and syntax in separate neural vectors and train a neural network with a specific structure or training objective to obtain disentangled vectors. 

In contrast, we leverage textual transformation to preserve the content or expression of a text while reducing another. Our approach is different in three ways. First, we focus on the discourse level instead of the sentence level, where the texts are longer and more complex. Second, we represent content and expression still in texts instead of neural vectors, which provides us with convenience for understanding and explaining. Finally, we use LLM with prompting techniques instead of training a model, which simplifies the usage and generalizes better.

\begin{figure*}[t]
    \centering
    \includegraphics[trim={0pt 0pt 0pt 0pt},clip,width=1.0\linewidth]{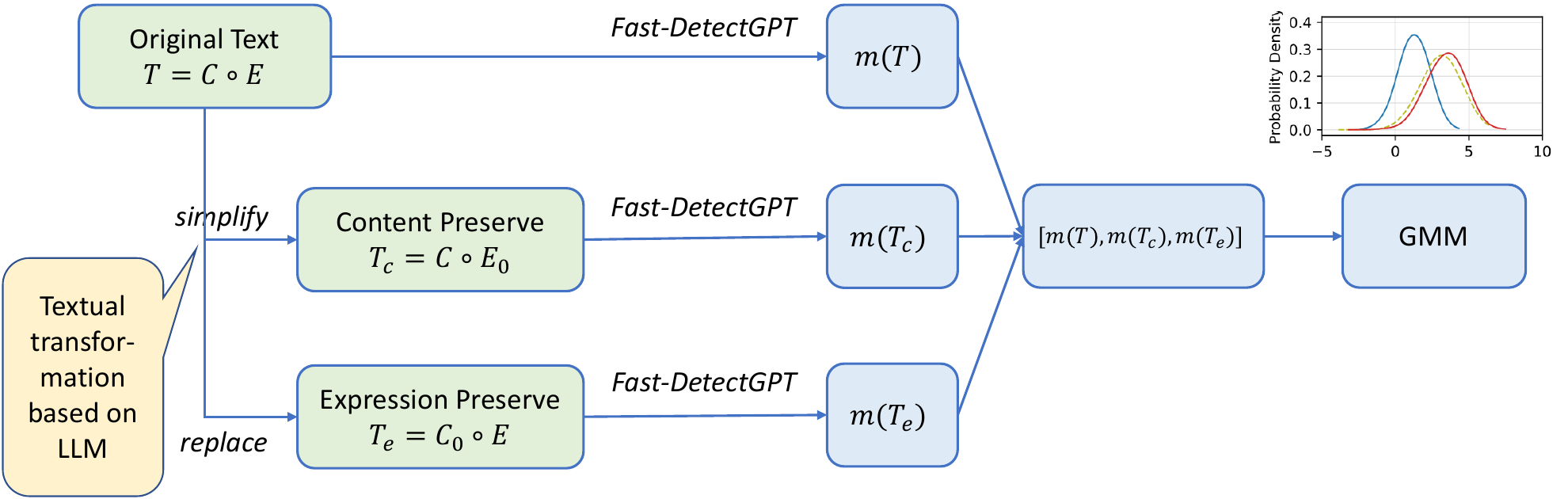}   
    \caption{Overview of the \emph{Triospect detection process}. The process consists of three steps: (i) \emph{Separating}, which transforms the original text into content-preserving and expression-preserving variants using an LLM; (ii) \emph{Measuring}, which computes detection scores using an existing detector (e.g., Fast-DetectGPT); and (iii) \emph{Classifying}, which performs binary classification with a Bayesian classifier based on a GMM.}
    \label{fig:detection_process}
\end{figure*}

\section{Triospect}
\label{sec:method}

The detection framework relies on measuring the content and expression of a text separately. We achieve the separation through textual transformations and measure them using existing detection metrics. Thus, the detection process involves \emph{separating}, \emph{measuring}, and \emph{classifying} steps, as illustrated in Figure \ref{fig:detection_process}.

\subsection{A Conceptual Model of Texts}

\revision{
The notions of content and expression are related to the classical distinction between semantics and grammar \cite{bickhard1993representational,dapretto1999form,chen2019multi,caucheteux2021disentangling,moro2001syntax}, but we define them at the discourse level rather than the sentence level. 
\begin{itemize}
    \item \textbf{Content} refers to the abstract, pre-linguistic meaning structure that can be expressed: it consists of semantic units (e.g., events, entities, propositions, imagery) and the relations among them (e.g., causal, temporal, contrastive). It also encompasses the underlying conceptual network in cognition -- intentions, emotions, viewpoints, and thematic structures -- which may exist independently of any particular linguistic form.
    \item \textbf{Expression} refers to the linguistic realization of this meaning structure: the concrete encoding of content through grammar, lexical choice, discourse organization, stylistic preference, pragmatic strategy, and other surface-level devices.
\end{itemize}

Psychological studies suggest that these two levels are partially separable in human text production \cite{kintsch1978toward,derose1997text,clark2007content,flower2016dynamics}. Writers typically engage in a content-planning stage, where conceptual and relational structures are organized, followed by an expression-design stage, where these structures are rendered into specific linguistic forms to suit communicative goals and audience expectations. Following this conceptualization, we assume that a text can be analytically decoupled into its content and expression. Conversely, given a well-specified content structure and a chosen mode of expression, a concrete text can be constructed as their realization.
}

% The concepts of content and expression are analogous to the concepts of semantics and syntax of texts \cite{bickhard1993representational,dapretto1999form,chen2019multi,caucheteux2021disentangling,moro2001syntax}, but they focus on the discourse level. Psychological studies show that the separation of content and expression occurs in the human text creation process \cite{kintsch1978toward,derose1997text,clark2007content,flower2016dynamics}, where a content planning stage is performed to plot the story and an expression design stage is followed to engage the readers. 
% We follow this conceptual model and assume that a text can be decoupled into its content and expression, and conversely, given a content and an expression, we can compose a text uniquely. 

Formally, we express a text $T$ as a composition of its content $C$ and expression $E$ as
\begin{equation}
  T=C \circ E,
\end{equation}
where $\circ$ denotes the composition operator.

\paragraph{AI-Generated Text Detection}
Theoretically, an AI-generated text $T_{M}$ can be expressed as a composition of an AI-planned content $C_M$ and an AI-designed expression $E_M$, while a human-written text $T_{H}$ can be expressed as a composition of a human-planned content $C_H$ and a human-designed expression $E_H$, as follows:
\begin{equation}
  T_{M}=C_{M} \circ E_{M}, \hspace{20pt}
  T_{H}=C_{H} \circ E_{H}.
\end{equation}

When we consider a humanizing attack by either an automatic tool or a human, the AI-generated texts are transformed to
\begin{equation}
  T_{M}^{'}=C_{M} \circ E_{M}^{'} \hspace{10pt} \text{ or } \hspace{10pt}
  T_{M}^{'}=C_{M} \circ E_{H},
\end{equation}
where the expression is converted into either an AI-altered version $E_{M}^{'}$ or a human style $E_{H}$.

Thus, AI-generated text detection tasks require the distinction of $\{T_M, T_M^{'}\}$ from $\{T_H\}$, which can be identified from either the content or the expression of the text if we can measure them separately.

\paragraph{The Challenge}
Separating the latent content and expression from observed texts is challenging because it requires a full understanding of what is said and how it is said. As a prototype, we utilize LLMs to approximate the function.

\begin{table*}[t]
    % \vspace{-0.2in}
    \centering\scriptsize
    \begin{NiceTabular}{p{0.05\linewidth}p{0.25\linewidth}p{0.55\linewidth}}
        \toprule
        \multicolumn{3}{p{0.95\linewidth}}{\textbf{Original Text ($T$)}: {\it Jane was walking through the park when she spotted a small puppy wandering aimlessly. She picked it up, read the collar tag, and took it to the address. The owner was grateful to have the puppy back.}} \\
        \midrule
        \multicolumn{3}{p{0.95\linewidth}}{\bf Content Preserving:} \\
        \addlinespace[0.5em]
        $f_1$ & \textbf{Prompt:} {\it Summarize the main ideas of the text in a clear and concise manner:$\textbackslash$n\{text\}} & \textbf{Output:} {\it Jane found a lost puppy in the park, read its collar tag, and returned it to its grateful owner.
        } \\
        \addlinespace[0.5em]
        $f_2$ & \textbf{Prompt:} {\it Outline the main points of the text to get a clear and concise picture of the content:$\textbackslash$n\{text\}} & \textbf{Output:} {\it 
        1. Jane was walking through the park.
        2. She noticed a small puppy wandering aimlessly.
        3. Jane found a collar tag on the puppy.
        4. She took the puppy to the address on the tag.
        5. The owner was grateful to Jane for returning the puppy.} \\
        \addlinespace[0.5em]
        $f_3$ & \textbf{Prompt:} {\it Simplify the text to make it clear and concise while preserving its meaning:$\textbackslash$n\{text\}} & \textbf{Output:} {\it Jane found a puppy in the park and returned it to its grateful owner after reading the collar tag.} \\
        \midrule
        \addlinespace[0.5em]
        \multicolumn{3}{p{0.95\linewidth}}{\bf Expression Preserving:} \\
        \addlinespace[0.5em]
        $f_4$ & \textbf{Prompt}: {\it Identify and list the representative language expressions used in the text:$\textbackslash$n\{text\}}  & \textbf{Output:} {\it 
        1. "walking through the park" - indicates Jane's action and setting.
        2. "spotted a small puppy" - describes Jane noticing the puppy.
        3. "wandering aimlessly" - characterizes the puppy's behavior.
        4. "picked it up" - describes Jane's action of picking up the puppy.
        } \\
        \addlinespace[0.5em]
        $f_5$ & \textbf{Prompt:} {\it Substitute the key ideas in the text with placeholders, ensuring the original tone, style, and language remain intact:$\textbackslash$n\{text\}} & \textbf{Output:} {\it [Person] was walking through [location] when [they] spotted a [subject] wandering [manner]. [They] picked it up, read the [identifier], and took it to the [destination]. The [recipient] was [reaction] to have the [subject] back.} \\
        \addlinespace[0.5em]
        $f_6$ & \textbf{Prompt:} {\it Replace the main points of the text with a generic topic while preserving the language expression:$\textbackslash$n\{text\}} & \textbf{Output:} {\it Alex was strolling through the garden when they noticed a tiny kitten meandering without direction. They scooped it up, checked the collar tag, and brought it to the listed location. The caretaker was thankful to have the kitten returned.} \\
        \bottomrule
    \end{NiceTabular}
    % \vspace{-0.2in}
    \caption{Textual transformations that we explore for \revision{approximately} preserving content and expression. The example demonstrates the function of each transformation.}
    \label{tab:decouple_content_language}
\end{table*}

\subsection{Separating Content and Expression}
\label{subsec:separating}
% 2) Measure content and expression -> basic idea (textual transformation + detection metric) -> measure content -> measure expression

\paragraph{Textual Transformation}
We investigate textual transformations using LLMs to preserve either the content or the expression of a text. LLMs have a strong ability to understand the content and expression of a text and manipulate them under the guidance of specific prompts. Thus, we can approximately achieve the goal by exploring suitable prompts.

Formally, given an original text $T$, \revision{an idea content-preserving function $f_c$ transforms the text into a new one $T_c$ as  
\begin{equation}
 T_c=f_c(T), \text{ that } T_c = C \circ E_0,
\end{equation}
and an idea expression-preserving function $f_e$ produces $T_e$ as
\begin{equation}
 T_e=f_e(T), \text{ that } T_e = C_0 \circ E,
\end{equation}
where the expression $E$ are content $C$ are diminished to an uniform $E_0$ and $C_0$.}

\paragraph{Content Preserving}
Using a specific prompt, we guide an LLM to convert a text $T$ into \revision{an uniform} expression while preserving its \revision{overall} meaning, producing \revision{an approximation $\hat{T}_c$ of the idea} content-preserving text $T_c$. Specifically, we investigate three intuitive techniques: summarizing, outlining, and simplifying, as illustrated in Table \ref{tab:decouple_content_language} (prompts $f_1$ to $f_3$). \emph{Summarizing} produces a summary of the main ideas of a text, \emph{outlining} generates a list of a text's main points, while \emph{simplifying} produces a simplified expression of the text. Basically, all these techniques are able to approximately preserve the content and reduce the stylistic expression. We explore these techniques and their alternatives empirically. 

\paragraph{Expression Preserving}
Similarly, we guide an LLM to convert a text $T$ to \revision{$\hat{T}_e$, which  approximates the idea expression-preserving text $T_e$} while reducing the influence of its content. We explore three approaches, including listing representative syntaxes of a text, substituting key ideas of a text with placeholders, and replacing the main points of a text with AI-generated ideas, as the prompts $f_4$ to $f_6$ in Table \ref{tab:decouple_content_language} illustrate. \emph{Listing} preserves typical expressions while reducing most ideas. \emph{Substituting} preserves the expression template while reducing the specific characters. \emph{Replacing} preserves the pattern of expression while substituting the content with AI-generated ideas $C_M$, regardless of whether the original content was AI-generated or human-planned.

\paragraph{Discussion}
All these transformations are not perfect. For example, even with a placeholder representing the characters, the expression template produced by substituting still contains concrete, meaningful verbs. Preserving the expression of a text while reducing its content is more challenging since expression naturally relies on specific content. Additionally, LLM transformations may also introduce some errors. However, we demonstrate that these imperfect representations of content and expression are still surprisingly beneficial for detection tasks.

\subsection{Measuring Content and Expression}

\revision{
Based on the transformations introduced in Section \ref{subsec:separating}, we obtain multiple rewritten variants of each text that approximately preserve either its content or its expression. We then apply AI-text detectors to these variants and use the resulting scores to measure content and expression signals.

The key intuition is that content-preserving transformations retain the main ideas of a text while reducing stylistic variation, whereas expression-preserving transformations retain stylistic patterns while weakening the original content. Therefore, detector scores on these transformed texts reflect different aspects of the original text.
}

We find that these latent aspects display identifiable token distributions for different sources, where existing detection metrics, typically zero-shot detection metrics, can be utilized to differentiate between the sources. Due to the statistical nature of these metrics, they are not sensitive to partial variances in the transformed texts, producing relatively stable measures.

\subsection{Binary Classifier}
% 3) Multidimensional detection algorithm -> measure vector -> Gaussian distribution -> estimate probability.
After converting a text into a measure vector \revision{$\bm{x}=[m(T),m(\hat{T}_c),m(\hat{T}_e)]$}, we project it into three-dimensional space. We model the distributions of samples in this space using two multivariate Gaussian Mixture Models: $\rho_H$ for human-written texts and $\rho_M$ for AI-generated texts. Each distribution is characterized by a group of mean vectors $\bm{\mu}_k$, representing the average measure values, and a group of covariance matrices $\bm{\Sigma}_k$, capturing the relationships and variability between measures. The $k\in [1..K]$ denotes the index of mixture components, where $K$ is the number of components.

Based on a number of development samples, we estimate the distribution parameters $\bm{\mu}_k^H$, $\bm{\Sigma}_k^H$, $\bm{\mu}_k^M$, and $\bm{\Sigma}_k^M$, obtaining two probability density functions
\begin{equation}
\small
  \rho_H(\bm{x}) = \sum_{k=1}^K \phi_k^H \mathcal{N}(\bm{\mu}_k^H, \bm{\Sigma}_k^H)
\end{equation}
and
\begin{equation}
\small
  \rho_M(\bm{x}) = \sum_{k=1}^K \phi_k^M \mathcal{N}(\bm{\mu}_k^M, \bm{\Sigma}_k^M),
\end{equation}
where $\phi_k^H$ and $\phi_k^M$ are the weights of mixture components, each totaling 1, and $\mathcal{N}(\cdot)$ is a single component of a multivariate Gaussian distribution. The component parameters $\bm{\mu}_k^H$ and $\bm{\mu}_k^M$ are $3$-dimensional mean vectors, and \revision{$\bm{\Sigma}_k^H$} and \revision{$\bm{\Sigma}_k^M$} are $3\times 3$ covariance matrices.
The probability density $\rho_H(\bm{x})$ and $\rho_M(\bm{x})$ describe the likelihood of a text with a measure vector $\bm{x}$ under the distributions of human-written and AI-generated texts, respectively.
% The numerator in each equation measures how far the measure $\bm{x}$ is from the mean $\bm{\mu}$, scaled by the covariance structure $\bm{\Sigma}$. Texts closer to the mean are more likely, while those farther away are less likely.

Our goal is to determine whether a given text is more likely to be AI-generated or human-written. To do this, we calculate the Bayesian posterior probability as
\begin{align}
\small
  p(M|\bm{x}) &= \frac{p(\bm{x}|M) p(M)}{p(\bm{x}|H) p(H) + p(\bm{x}|M) p(M)} \\
  &= \frac{\rho_M(\bm{x})}{\lambda \cdot \rho_H(\bm{x}) + \rho_M(\bm{x})},    
\end{align}
where $\lambda=p(H)/p(M)$ denotes the ratio of the prior probabilities. 
For balanced classes, $\lambda$ is set to 1, whereas for imbalanced classes, it represents the ratio of human samples to machine samples. To arrive at a final decision, a probability threshold is required, which should be determined based on the specific application.

\paragraph{Discussion}
Empirically, we observe that \emph{a single Gaussian component is typically sufficient} across different scenarios, with extra components offering only marginal improvements. This may be due to two factors. First, the detection metric is based on token averages. By the Central Limit Theorem, the mean of many random variables tends to approximate a normal distribution, regardless of their original forms. Second, the detection decision is only sensitive to the density functions near the boundary where $\rho_H$ and $\rho_M$ have close values. The decision boundary usually falls within low-density regions, influencing only a small subset of samples. Moreover, in these regions, the density estimated by a single component is often not substantially different from that estimated with multiple components.

\begin{table}[t]
    % \vspace{-0.2in}
    \centering\scriptsize
    \begin{tabular}{@{}lcc@{}}
        \toprule
        \bf Domain & \bf Avg. Words & \bf Dev / Test \\
        \midrule
        Student Essay \cite{scott2024asap} & 241 & 2K / 2K \\
        ArXiv Intro \cite{arxiv2024} & 410 & 2K / 2K \\
        Creative Writing \cite{fan2018hierarchical}  & 345 & 2K / 2K \\
        CC News \cite{Hamborg2017} & 148 & 2K / 2K \\
        \midrule
        % CC News & Chinese & 590 chars & 2K / 2K \\
        CC News (French) & 258 & 2K / 2K \\
        CC News (Spanish) & 285 & 2K / 2K \\
        CC News (Arabic) & 152 & 2K / 2K \\
        \bottomrule
    \end{tabular}
    \caption{Humanize-16K and a multilingual patch.}
    \label{tab:datasets}
\end{table}

\section{Benchmark of Humanizing Attack}
% 1) Humanizing attack: human written -> model generated -> humanizing by human-editing, AI tools, LLM (four domains, six models + r1, decoding strategies)
We create a dataset \emph{Humanize-16K} to mimic a real-world situation influenced by human editing and humanizing tools, focusing on a difficult scenario not addressed by current benchmarks.
Our dataset covers 6 humanizing approaches, 4 domains, 6 text generation models, and various decoding strategies. We describe the detailed construction of the dataset in Appendix \ref{app:datasets} and summarize it as follows.
% Detailed statistics are provided in Table .

\revision{
\subsection{Construction of Humanize-16K}
}

% We collect English human-written texts $T_H$ from the Automated Student Assessment Prize (ASAP) 2.0 dataset \cite{scott2024asap} for \emph{student essays}, from arXiv \cite{arxiv2024} for \emph{paper introductions}, from WritingPrompts \cite{fan2018hierarchical} for \emph{story writings}, and from Common Crawl \cite{Hamborg2017} for \emph{news articles}, with 2,000 human samples for each domain.

% We produce corresponding AI-generated texts $T_M$ using LLMs including \emph{gpt-3.5-turbo}, \emph{gpt-4o}, \emph{claude-3.5-sonnet}, \emph{gemini-1.5-pro}, \emph{llama-3.3-70b-instruct}, and \emph{qwen-2.5-72b-instruct}. We generate the texts based on human-written titles for arXiv papers and CC News, and human-written prompts for student Essays and WritingPrompts, by randomly selecting half of the human samples. The process produces 1,000 machine samples for each domain.

% We humanize AI-generated texts by \emph{human editing}, \emph{commercial humanizing tools}, and \emph{simulated humanizing process using LLMs}. We hire five professional annotators for human editing, producing 250 human-edited samples. We select bypassgpt.ai, humbot.ai, undetectable.ai as representatives of commercial humanizing tools. We further simulate the humanizing process using LLM, including \emph{diversifying} lexicon and syntax and \emph{mimicking} human-writing styles. This step produces 1,000 humanized samples per domain.

As the domains listed in Table \ref{tab:datasets}, we collect 2,000 human samples $T_H$ from each domain. For a random half $T_{H0}$ of the human samples, we produce corresponding AI-generated texts $T_M$ using 6 LLMs, including \emph{gpt-3.5-turbo}, \emph{gpt-4o}, \emph{claude-3.5-sonnet}, \emph{gemini-1.5-pro}, \emph{llama-3.3-70b-instruct}, and \emph{qwen-2.5-72b-instruct}. We further humanize these generated texts using \emph{human editing}, \emph{commercial AI tools}, and \emph{simulated humanizing process}, producing 1,000 humanized samples $T_M^{'}$ per domain.
Consequently, we obtain an equal number of human-written texts and AI-generated/humanized texts, resulting in a total of 16,000 samples.

We configure the task into three subsets: the \emph{before-attack} subset for classifying between $\{T_{H0}\}$ and $\{T_M\}$, the \emph{after-attack} subset for classifying between $\{T_{H0}\}$ and $\{T_M^{'}\}$, and the full set \emph{mix} for classifying between $\{T_H\}$ and $\{T_M,T_M^{'}\}$ \revision{, where the complete human set $T_H$ is used to balance the two classes}. We randomly split the samples into a development set and a test set of equal size for each.

Additionally, we create a \emph{multilingual patch} by collecting \emph{French}, \emph{Spanish}, and \emph{Arabic} news from Common Crawl \cite{Hamborg2017}, and follow the same process to produce 4,000 samples per language.

In the construction process, we hire human annotators for editing and quality checks of the generated texts. Additionally, we call the API for the commercial AI tools. Totally, they cost about \$2500. 

\revision{
\subsection{Humanizing AI-Generated Texts}
\label{app:humanizing}

We humanize AI-generated text via human editing, commercial tools, and automated LLM-based simulations.

\subsubsection{Human Editing}
\label{app:human_editing}

We employ five annotators from a specialized annotation company, consisting of three individuals with professional expertise in English and two with a background in computer science. The team includes three women and two men, aged 22 to 41 years, all of whom have Chinese as their native language. Each annotator is responsible for revising 50 AI-generated texts, resulting in a total of 250 human-edited samples. 

The editing process is carried out at three levels: word, sentence, and paragraph. At the word level, synonyms are used to replace existing words; at the sentence level, syntax alterations are made; and at the paragraph level, the logical flow of sentences is reorganized. Annotators are asked to apply these three types of edits in equal proportion, ensuring that more than 50\% of the original content is modified. In addition, a separate annotator reviews 10\% of the texts to verify that the edits preserve the original meaning while ensuring that the revised texts remain fluent and comprehensible.

It costs about \$2,000 for human editing.

\subsubsection{Commercial AI Tools}
We use three commercial AI tools, including \emph{humbot.ai}, \emph{bypassgpt.ai}, and \emph{undetectable.ai}, from Table \ref{tab:list_humanizing_system} in Appendix. 
We call these tools through their API, producing 765 humanized documents. The API costs about \$500 in total.

\subsubsection{Simulation by LLMs}

We use the following prompts to humanize AI-generated text by LLMs.

\paragraph{Prompt for Diversifying:} ``{\it Revise the text to enrich its linguistic diversity, employing varied sentence structures, synonyms, and stylistic nuances, while preserving the original meaning:$\backslash$n\{generation\}}''

\paragraph{Prompt for Mimicking:} ``{\it Rewrite the text using the same language style, tone, and expression as the reference text. Focus on capturing the unique vocabulary, sentence structure, and stylistic elements evident in the reference:$\backslash$n\{generation\}$\backslash$n$\backslash$n\# Reference Text:$\backslash$n\{reference\}}''

Diversifying enriches expression while preserving core ideas. Mimicking creates more human-like text by copying style and structure but may introduce fabricated content and alter the original text.
}

\begin{figure*}[t]
\begin{minipage}{0.48\linewidth}
    \centering
    \includegraphics[trim={0pt 0pt 0pt 0pt},clip,width=1.0\linewidth]{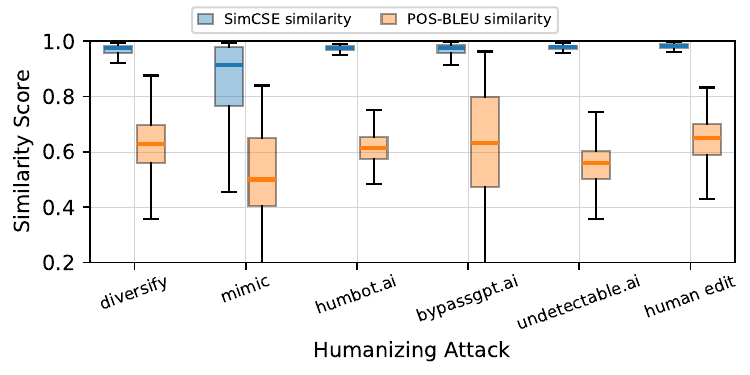}   
    \captionof{figure}{Semantic and grammar similarities per humanizing attack.}
    \label{fig:attack_similarity_boxplot}
\end{minipage}
\hfill\hspace{6pt}
\begin{minipage}{0.48\linewidth}
    \centering
    \includegraphics[trim={0pt 20pt 0pt 0pt},clip,width=1.0\linewidth]{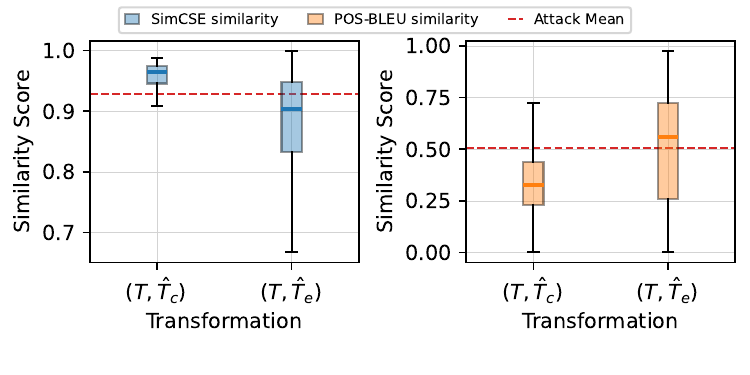}   
    \captionof{figure}{Semantic and grammar similarities for content-preserving ($f_3$) and expression-preserving ($f_6$) transformations.}
    \label{fig:transform_similarity_boxplot}
\end{minipage}
\end{figure*}

\section{Experimental Settings}

\subsection{Datasets}
\label{sec:dataset}

% 2) Adversarial attacks: eight domains, 11 models, 11 adversarial attacks, decoding strategies

We use our Humanize-16K benchmark and the existing benchmark RAID \cite{dugan2024raid} as representatives of wild scenarios under attack. Humanize-16K covers 6 humanizing attacks, 4 domains, and 6 models. RAID covers 11 adversarial attacks, 8 domains, 11 models, and various decoding strategies, where we sample 4K for testing and another 4K for development.

\subsection{Detectors}
\paragraph{Baselines}
We consider classical and state-of-the-art \emph{zero-shot detectors}, which generally leverage pre-trained LLMs to compute a detection metric as an indicator of AI-generated text. These metrics can also be used in the Triospect detection framework. Specifically, we take log-perplexity, log-rank, LRR \cite{su2023detectllm}, Fast-Detect \cite{bao2024fast}, Binoculars \cite{hans2024spotting}, Glimpse \cite{bao2025glimpse}, and Raidar \cite{maoraidar2024} as representatives. We use falcon-7B as the scoring model for the first three detectors, falcon-7B/falcon-7B-instruct as the scoring models for Fast-Detect and Binoculars. Raidar extracts 56-dimensional features from multiple rewritten texts and trains a binary classifier.

We also consider \emph{supervised detectors}, using RADAR \cite{hu2023radar}, RoBERTa (ChatGPT) \cite{guo2023hc3}, and ImBD (imitate before detect) \cite{chen2025imitate} as representatives. Typically, RADAR is trained under the concept of adversarial learning, which can resist paraphrasing attacks. ImBD uses a fine-tuned gpt-neo-2.7B on AI-rewritten texts, which is optimized for detecting AI texts refined by a humanizing tool.

\paragraph{Our Detector}
We present \emph{Triospect} combined with a specific existing detection metric. In our main experiments, we utilize $f_3(T)$ to generate \revision{$\hat{T}_c$} and $f_6(T)$ to generate \revision{$\hat{T}_e$}, employing \emph{Qwen3-4B} (with max\_tokens = 200) under non-thinking mode as the transformation model. We use a single mixture component ($K=1$) in the main experiments and compare this setting with alternatives in the ablation study.

\subsection{Metrics}
We use \emph{AUROC}, the area under the receiver operating characteristic curve, as the major metric to measure the quality of the classifiers. We report \emph{ACC} (accuracy) with the best threshold found by maximizing Youden's J statistics \cite{youden1950index}. We also report \emph{TPR01}, a true positive rate at a false positive rate of 1\%, for reference. We do significance test with McNemar's Test \cite{mcnemar1947note} for ACC, and Bootstrap Resampling \cite{dwivedi2017analysis} for TPR01, with a $p$-value less than 0.01. 
We run all the experiments on a machine with one Tesla A100 GPU, which takes about six hours.

\begin{table*}[t]
    % \vspace{-0.2in}
    \centering\scriptsize
    \begin{tabular}{lccccccccc}
        \toprule
        \multirow{2}{*}{\bf Detector}& \multicolumn{3}{c}{\bf AUROC} & \multicolumn{3}{c}{\bf ACC} & \multicolumn{3}{c}{\bf TPR01} \\
        \cmidrule(r){2-4}\cmidrule(r){5-7}\cmidrule(r){8-10}
        & \bf Before & \bf After & \bf Mix & \bf Before & \bf After & \bf Mix & \bf Before & \bf After & \bf Mix \\
        \midrule
        RoBERTa(ChatGPT) \cite{guo2023hc3}  & 0.664 & 0.560 & 0.563 & 61\% & 60\% & 61\% & 8\% & 0\% & 5\% \\
        RADAR \cite{hu2023radar} & 0.838 & 0.729 & 0.787 & 76\% & 67\% & 70\% & 5\% & 3\% & 6\% \\
        Log-Perplexity & 0.764 & 0.647 & 0.549 & 72\% & 59\% & 56\% & 0\% & 4\% & 0\% \\
        Log-Rank & 0.779 & 0.658 & 0.552 & 73\% & 60\% & 58\% & 0\% & 6\% & 0\% \\
        LRR \cite{su2023detectllm} & 0.820 & 0.656 & 0.577 & 76\% & 62\% & 61\% & 25\% & 2\% & 11\% \\
        Raidar \cite{maoraidar2024} & 0.929 & 0.911* & 0.880 & 84\% & 76\% & 80\% & 77\% & 41\% & 19\% \\
        \hdashline
        Binoculars \cite{hans2024spotting} & 0.916 & 0.629 & 0.772 & 91\% & 67\% & 79\% & 83\% & 30\% & 56\% \\
        \it Triospect (Binoculars) & \bf 0.967 & \bf 0.848 & \bf 0.908 & \bf 93\%* & \bf 78\% & \bf 84\% & \bf 85\%* & \bf 35\% & \bf 59\% \\
        \hdashline
        ImBD \cite{chen2025imitate} & 0.962 & 0.821 & 0.891 & 93\% & 74\% & 83\% & 85\% & 34\% & 59\% \\
        \it Triospect (ImBD) & 0.968* & \bf 0.879  & \bf 0.925* & 92\% & \bf 80\%* & \bf 85\%* & 85\% & \bf 41\% & 59\% \\
        \hdashline
        Fast-Detect \cite{bao2024fast} & 0.913 & 0.627 & 0.770 & 90\% & 68\% & 78\% & 81\% & 31\% & 55\% \\
        \multirow{2}{*}{\it Triospect (Fast-Detect)} & \bf 0.960 & \bf 0.850 & \bf 0.901 & \bf 92\% & \bf 78\% & \bf 84\% & \bf 84\% & \bf 44\%* & \bf 63\%* \\
        & \diff{(+0.047)} & \diff{(+0.223)} & \diff{(+0.131)} & \diff{(+2\%)} & \diff{(+10\%)} & \diff{(+6\%)} & \diff{(+3\%)} & \diff{(+13\%)} & \diff{(+8\%)} \\
        \bottomrule
    \end{tabular}
    \caption{\emph{Main results before and after humanizing attacks} evaluated on Humanize-16K. The significantly improved scores among each group are marked in \textbf{bold} and the best scores are marked with `*'. }
    \label{tab:main_results_humanize16k}
\end{table*}

\begin{table*}[t]
    \vspace{-0.1in}
    \setlength{\tabcolsep}{4pt}
    \centering\scriptsize
    \begin{tabular}{l@{\hspace{15pt}}cccccccc@{\hspace{15pt}}ccc}
        \toprule
        \multirow{2}{*}{\bf Detector} & \multicolumn{8}{c}{\bf AUROC per Domain of RAID} & \multicolumn{3}{c}{\bf Mixture of Domains} \\
        \cmidrule(r){2-9}\cmidrule{10-12}        
         & \bf News & \bf Books & \bf Wiki & \bf Abstracts & \bf Reddit & \bf Recipes & \bf Poetry & \bf Reviews  & \bf AUROC & \bf ACC & \bf TPR01 \\
        \midrule
        Binoculars & 0.768 & 0.850 & 0.804 & 0.826 & 0.811 & 0.759 & 0.826 & 0.812 & 0.807 & 78\% & 46\% \\
        \it Triospect (Binoculars) & \bf 0.887* & \bf 0.930 & \bf 0.881 & \bf 0.913* & \bf 0.898* & \bf 0.897* & \bf 0.920* & \bf 0.879 & \bf 0.901* & \bf 84\%* & \bf 59\% \\
        \hdashline
        ImBD & 0.783 & 0.875 & 0.804 & 0.806 & 0.825 & 0.699 & 0.823 & 0.840 & 0.804 & 76\% & 41\% \\
        \it Triospect (ImBD) & \bf 0.857 & \bf 0.954* & \bf 0.876 & \bf 0.865 & \bf 0.866 & \bf 0.843 & \bf 0.874 & \bf 0.895* & \bf 0.873 & \bf 82\% & \bf 56\% \\
        \hdashline
        Fast-Detect & 0.761 & 0.845 & 0.803 & 0.821 & 0.794 & 0.749 & 0.818 & 0.810 & 0.800 & 77\% & 39\% \\
        \multirow{2}{*}{\it Triospect (Fast-Detect)} & \bf 0.878 & \bf 0.925 & \bf 0.891* & \bf 0.897 & \bf 0.871 & \bf 0.895 & \bf 0.897 & \bf 0.880 & \bf 0.891 & \bf 83\% & \bf 61\%* \\
        & \diff{(+0.117)} & \diff{(+0.080)} & \diff{(+0.088)} & \diff{(+0.076)} & \diff{(+0.077)} & \diff{(+0.146)} & \diff{(+0.079)} & \diff{(+0.070)} & \diff{(+0.091)} & \diff{(+6\%)} & \diff{(+22\%)} \\
        \bottomrule
    \end{tabular}
    \vspace{-0.1in}
    \caption{\emph{Main results under adversarial attacks}, evaluated on RAID benchmark.}
    \label{tab:main_results_raid}
\end{table*}

\section{Experiments}

\revision{
\subsection{Necessity of Content and Expression Measures}
\label{sec:md_view}

Triospect assumes that \emph{AI-generated and human-written texts differ in content and expression, while content remains relatively stable under attacks}. To test this hypothesis, we use two complementary similarity metrics.

\emph{Content similarity} is measured via SimCSE \cite{gao2021simcse} with RoBERTa-large \cite{liu2019roberta}, which computes cosine similarity between sentence embeddings to capture semantic consistency. 
\emph{Expression similarity} uses POS-BLEU, which combines POS n-grams \cite{koppel2009computational} that reflect syntactic patterns with BLEU \cite{papineni2002bleu} to quantify surface-form overlap. 
These metrics allow separate analysis of meaning and form under attacks or transformations.

\paragraph{Effect of Attacks}
We first analyze how humanizing attacks modify texts. 
As shown in Figure~\ref{fig:attack_similarity_boxplot}, attacks mainly alter \emph{surface expression} rather than \emph{underlying content}. 
Specifically, the coefficient of variation (std/mean) of POS-BLEU scores reaches 23.1\%, whereas SimCSE scores vary by only 6.1\%. 
The much smaller variance of SimCSE indicates that semantic content remains largely stable under attack, while grammatical realization changes substantially. It suggests that \emph{content-based signals are inherently more robust under adversarial rewriting}.

\paragraph{Content- and Expression-Preserving Transformations}
We next evaluate whether the proposed transformations successfully isolate the two aspects of text. As shown in Figure~\ref{fig:transform_similarity_boxplot}, the \emph{content-preserving transformation} ($T \rightarrow \hat{T}_c$) maintains high SimCSE similarity but exhibit low POS-BLEU similarity, indicating that semantic content is preserved while surface expression changes. This confirms that the transformation effectively perturbs expression without altering meaning.

The \emph{expression-preserving transformation} ($T \rightarrow \hat{T}_e$) exhibits higher POS-BLEU similarity but lower SimCSE similarity, demonstrating that expression is preserved while semantics change. 
However, the similarity scores show larger variance than in the content-preserving case, suggesting that preserving expression while altering meaning is intrinsically more difficult to control.

\paragraph{Complementarity of Content and Expression Measures}
Finally, we examine whether AI and human texts differ under the proposed measures. As shown in Figure~\ref{fig:single_vs_2d} (middle), AI-generated and human-written texts exhibit distinguishable distributions in the content measure, which complements the differences observed in expression as illustrated by the skewed linear decision boundary in Figure~\ref{fig:single_vs_2d} (right). 
These results indicate that jointly modeling content and expression is necessary for reliable AI-generated text identification.
}

\subsection{Main Results}
\label{sec:main_results}
% 1) main results: resist to humanizing attack and adversarial attack
We first evaluate the detectors on their ability to mitigate humanizing attacks. As Table \ref{tab:main_results_humanize16k} shows, all detectors experience a significant performance drop after the attacks. Take Fast-Detect as an example. The attacks reduce the AUROC from 0.913 to 0.627. Compared to the baseline, Triospect significantly mitigates the impact of the attacks, increasing the AUROC from 0.627 to 0.850. Surprisingly, Triospect also enhances the AUROC on texts before the attacks. The effects on multiple baselines are consistent. Typically, upon the strong baseline ImBD, which has been optimized for AI-rewritten texts, the Triospect detector still achieves significant improvements in AUROC and ACC.

We further compare the detectors on their ability to mitigate adversarial attacks using the RAID benchmark. As shown in Table \ref{tab:main_results_raid}, the Triospect detectors outperform the baselines by even larger margins. Triospect enhances the AUROC by 9.1\% and TPR01 by 22\% for Fast-Detect, and the AUROC by 9.4\% and TPR01 by 13\% for Binoculars. 

\begin{table}[t]
    % \vspace{-0.2in}
    \centering\scriptsize
    \begin{tabular}{@{}l@{\hspace{5pt}}c@{\hspace{5pt}}c@{\hspace{5pt}}c@{\hspace{5pt}}c@{\hspace{5pt}}c@{}}
        \toprule
        \multirow{2}{*}{\bf Method} & \multicolumn{2}{c}{\bf Humanize-16K} & \multicolumn{2}{c}{\bf RAID} & \bf Time \\
        % \cmidrule(r){2-3}\cmidrule{4-5}
        & \bf AUROC & \bf TPR01 & \bf AUROC & \bf TPR01 & \bf /Sample \\
        \midrule
        Raidar & 0.877 & 19\% & 0.773 & 17\% & 19.5s \\
        Fast-Detect & 0.770 & 55\% & 0.800 & 39\% & \bf 0.2s \\
        % \hdashline
        \multicolumn{3}{@{}l}{\it Triospect (Fast-Detect) using Qwen3-4B} \\
        \hspace{10pt} max\_tokens=25 & 0.870 & 62\% & 0.868 & 42\% & 0.5s \\
        \hspace{10pt} max\_tokens=50 & 0.875 & 62\% & 0.878 & 50\% & 0.7s \\
        \hspace{10pt} max\_tokens=100 & 0.890 & 63\% & 0.886 & 56\% & 1.3s \\
        \hspace{10pt} max\_tokens=200 & 0.901 & 63\% & \bf 0.893 & \bf 61\% & 2.1s \\
        % \hdashline
        \multicolumn{3}{@{}l}{\it Triospect (Fast-Detect) using GPT-4o} \\
                      & \bf 0.909 & \bf 65\% & 0.891 & 59\% & 6.5s \\
        \bottomrule
    \end{tabular}
    \caption{\emph{Detection efficiency}, where the detection accuracy and speed can be balanced by the number of tokens generated.}
    \label{tab:ablation_efficiency}
\end{table}

The results on the two datasets demonstrate the effectiveness of Triospect in addressing attacks. Although they are training-free, they achieve better detection performance than existing supervised and zero-shot detectors. Additionally, the improvements are consistent across domains and evaluation metrics.

\paragraph{Efficiency}
As shown in Table \ref{tab:ablation_efficiency}, Triospect requires about 2.1 seconds per sample with max\_tokens=200 and 0.5 seconds with max\_tokens=25. For a max\_tokens of 25, its runtime is comparable to Fast-Detect while still achieving a notable improvement in AUROC and TPR01. In practice, we could strike a balance between detection accuracy and computational efficiency by constraining the maximum number of generated tokens in the transformations.
% Among these settings, the Triospect detectors run much faster than rewrite-detect Raidar.

\subsection{Ablation Study}
\label{sec:ablation}
% 2) ablation study: rely on LLM (model, prompts, decoding strategy, size of dev, Gaussian mixture) and metric (trained detecor) by a subset
% We sample 1/4 of the full set across the four domains for the ablation, with 2,000 samples for dev and 2,000 samples for test.

\paragraph{Feature Contribution}
Triospect incorporates two extra features: content and expression measures. Our findings indicate that the content feature drives most of the performance gains. Specifically, the feature set $[m(T), m(T_c)]$ yields an AUROC of 0.898 for Fast-Detect on Humanize-16 mix, nearly matching the 0.901 AUROC obtained with the full feature set. However, it may vary against different attacks. In the context of RAID, the expression measure has a marginally greater impact, which increases the AUROC from 0.880 to 0.891.

\begin{figure}[t]
% \vspace{-0.1in}
\begin{minipage}{0.40\linewidth}
    \centering
    \includegraphics[trim={0pt 0pt 0pt 0pt},clip,width=1.0\linewidth]{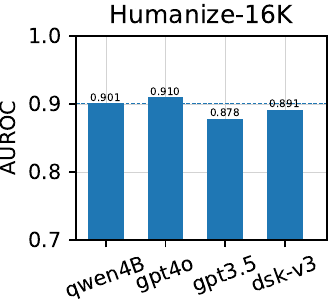}   
    \captionof{figure}{Transformation LLMs.}
    \label{fig:ablation_llm_bartchart}
\end{minipage}
\hfill\hspace{6pt}
\begin{minipage}{0.55\linewidth}
    \centering
    \includegraphics[trim={0pt 0pt 0pt 0pt},clip,width=1.0\linewidth]{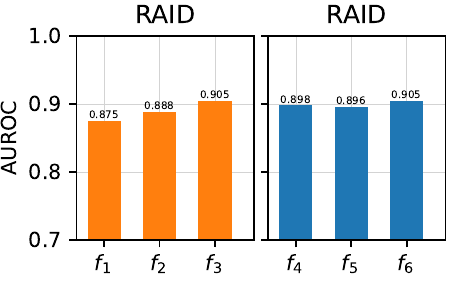}   
    \captionof{figure}{Transformation prompts.}
    \label{fig:ablation_prompt_bartchart}
\end{minipage}
% \vspace{-0.1in}
\end{figure}

\begin{figure*}[t]
% \vspace{-0.2in}
\begin{minipage}{0.33\linewidth}
    \centering
    \includegraphics[trim={0pt 0pt 0pt 0pt},clip,width=1.0\linewidth]{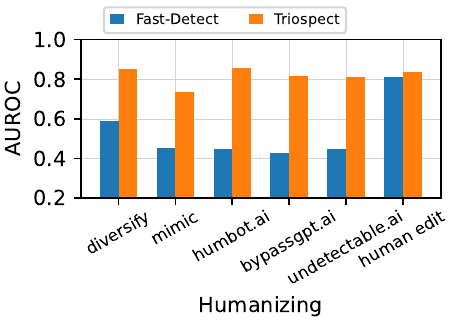}   
    \caption{Against humanizing attacks (Humanize-16K).}
    \label{fig:main_hart_analysis_barchart}
\end{minipage}
\hfill\hspace{5pt}
\begin{minipage}{0.66\linewidth}
    \centering
    \includegraphics[trim={0pt 0pt 0pt 0pt},clip,width=1.0\linewidth]{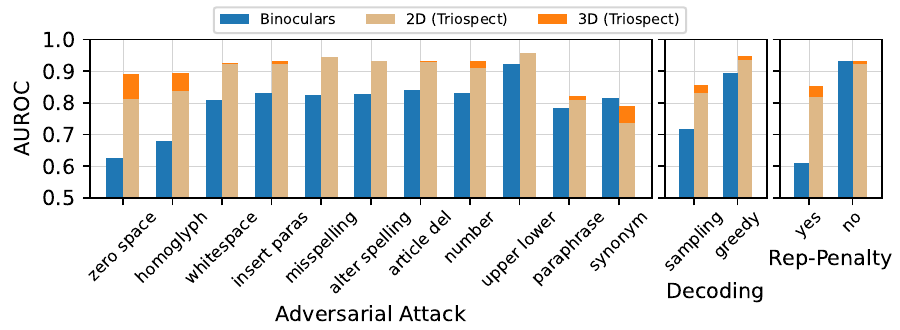}   
    \caption{Against adversarial attacks and deal with various decoding strategies (RAID).}
    \label{fig:main_raid_analysis_barchart}
\end{minipage}
\end{figure*}

\paragraph{Textual Transformation}
The choice of \emph{LLM} for transformation affects detection performance. As shown in Figure \ref{fig:ablation_llm_bartchart}, Triospect (Fast-Detect) achieves higher AUROC with the stronger gpt-4o but lower with gpt-3.5 and deepseek-v3. We also examine sampling strategies and find that setting top-$p$ with $0.6$ improves AUROC by about 1\%, whereas greedy decoding reduces it by roughly 3\%.

The design of transformation \emph{prompts} also impacts performance. Figure \ref{fig:ablation_prompt_bartchart} shows that Triospect (Binoculars) with $f_3$ outperforms $f_1$ and $f_2$, while using $f_6$ outperforms $f_4$ and $f_5$. We further test alternative versions of $f_3$ and $f_6$ by asking gpt-4o to rewrite each prompt five times. Results show only marginal variation across alternatives, with AUROC fluctuations within $\pm 0.009$.

Finally, LLM transformations can introduce \emph{errors}, potentially affecting detection accuracy. To assess this, we analyze prediction consistency across multiple generations of the same input text. Since sampling introduces varying errors, we estimate their effects on prediction probabilities. Results indicate that these errors lead to prediction scores with a standard deviation of 0.012, which is negligible for detection.

\paragraph{Detection Metric}
We also try other detection metrics, such as Log-Perplexity and Log-Rank, obtaining consistent improvements. We further try trained detectors such as RoBERTa (ChatGPT) and RADAR, using their predictive probabilities as detection metrics. However, empirical results show that trained detectors are ineffective in measuring transformed texts $T_c$ and $T_e$. This is probably because the trained detectors are not familiar with these text styles.

\paragraph{Density Estimation}
Triospect estimates its parameters using a \emph{dev set}. In practice, just 20 sample pairs are sufficient for full performance with a single Gaussian component, achieving an AUROC of 0.910 on Humanize-16K for Triospect (Fast-Detect), which is even slightly better than the 0.901 AUROC obtained with the full dev set.

We evaluate both single and multiple Gaussian components. Results show that using 2 to 5 components offers no clear advantage over a single component, with AUROC varying within $\pm0.005$. Notably, the single-component model requires fewer samples for stable estimation.

% We further test the parameter estimation in an `out-of-domain' setting, where for testing of each domain we exclude the `in-domain' samples from the dev set. Experiments show that the parameters are not sensitive to the setting, resulting in a marginal performance drop by an average AUROC of 0.007 for Triospect (Fast-Detect) on our dataset.

\subsection{Analysis of Robustness}
\label{sec:robustness}

\paragraph{Against Humanizing Attacks}
The analysis in Figure \ref{fig:main_hart_analysis_barchart} indicates that AI tools are the major source of threat, unlike human editing, which the baseline detector identifies with high accuracy. The baseline detector struggles significantly, often failing against commercial AI tools (e.g., humbot.ai, bypassgpt.ai, undetectable.ai). In contrast, Triospect shows marked improvement in AUROCs, highlighting its resilience against current commercial humanizing AI.

\begin{figure}[t]
    \centering
    \includegraphics[trim={0pt 0pt 0pt 0pt},clip,width=0.6\linewidth]{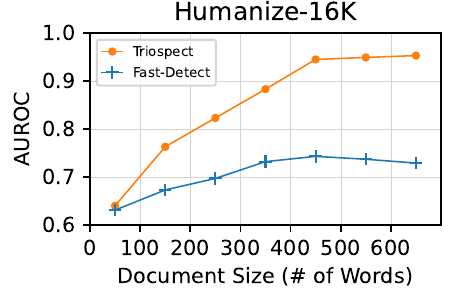}   
    \caption{AUROC on Humanize-16K mix, grouped by text lengths.}
    \label{fig:hart_length_linechart}
\end{figure}

\paragraph{Against Adversarial Attacks and Decoding Strategies}
As Figure \ref{fig:main_raid_analysis_barchart} demonstrates, Triospect detectors are robust against adversarial attacks and decoding strategies. Triospect achieves significant improvements in all categories except `synonym' and `non-repetition-penalty'. Among them, the expression measure mainly contributes to zero-space, homoglyph, and synonym attacks, as shown by the comparison between the 2D $[m(T),m(T_c)]$ and 3D $[m(T),m(T_c),m(T_e)]$.

\paragraph{Over Different Lengths}
We categorize the samples into buckets such as $[0, 100)$, $[100, 200)$, $[200, 300)$, etc., and compare Triospect with the baseline in Figure \ref{fig:hart_length_linechart}. Generally, the longer the texts are, the greater the improvement will be, where Triospect is especially beneficial for long texts.

\begin{table}[t]
    % \vspace{-0.2in}
    \setlength{\tabcolsep}{1.5pt} 
    \centering\scriptsize
    \begin{tabular}{@{}lcccccc@{}}
        \toprule
        \bf Detector & \bf qwen2.5 & \bf gemini1.5 & \bf llama3.3 & \bf gpt3.5 & \bf claude3.5 & \bf gpt4o \\
        \midrule
        Fast-Detect & 0.750 & 0.756 & 0.775 & 0.716 & 0.752 & 0.490 \\
        Triospect & \bf 0.879 & \bf 0.888 & \bf 0.888 & \bf 0.878 & \bf 0.850 & \bf 0.810 \\
        \bottomrule
    \end{tabular}
    \caption{\emph{AUROC per source LLM} evaluated on Humanize-16K mix.}
    \label{tab:results_per_llm}
\end{table}

\paragraph{Across Source Models}
We compared detectors on texts generated by different source models. We conduct both `in-domain' and `out-of-domain' settings, where we use dev samples including the model for the `in-domain' setting and those excluding the model for the `out-of-domain' setting. Experiments show that the scores are almost identical in the two settings. As Table \ref{tab:results_per_llm} shows, Triospect outperforms the baseline in all categories. The improvements are especially significant for stronger models such as gpt-4o.

\begin{table}[t]
    % \vspace{-0.2in}
    % \setlength{\tabcolsep}{3pt} 
    \centering\scriptsize
    \begin{tabular}{lcccccc}
        \toprule
        \multirow{2}{*}{\bf Detector} & \bf French & \bf Spanish & \bf Arabic \\
         & \bf in / out & \bf in / out & \bf in / out \\
        \midrule
        Fast-Detect & 0.773 & 0.696 &  0.465 \\
        Triospect & \bf 0.841 / 0.841 &\bf 0.791 / 0.790 & \bf 0.616 / 0.614 \\
        \bottomrule
    \end{tabular}
    \vspace{-0.1in}
    \caption{\emph{AUROC per language} evaluated on multilingual patch, where `{\it (in)}' denotes `in-domain' setting  and `{\it (out)}' denotes `out-of-domain' setting.}
    \label{tab:results_per_language}
\end{table}

\begin{table}[t]
    % \vspace{-0.2in}
    \setlength{\tabcolsep}{1.5pt} 
    \centering\scriptsize
    \begin{tabular}{@{}lcccccc@{}}
        \toprule
        \multirow{2}{*}{\bf Detector} & \bf Writing & \bf News & \bf Essay & \bf ArXiv \\
         & \bf in / out & \bf in / out & \bf in / out & \bf in / out \\        
        \midrule
        Fast-Detect & 0.730 & 0.729& 0.862 & 0.755  \\
        Triospect & \bf 0.915 / 0.913 & \bf 0.883 / 0.879 & \bf 0.874 / 0.866 & \bf 0.934 / 0.933  \\
        \bottomrule
    \end{tabular}
    \caption{\emph{AUROC per domain} evaluated on Humanize-16K.}
    \label{tab:results_per_domain}
\end{table}

\paragraph{Across Languages}
We assess detector performance on French, Spanish, and Arabic news data under two settings: `in-domain' with dev samples from both English and the language, and `out-of-domain' with dev samples from English only. As Table \ref{tab:results_per_language} shows, Triospect outperforms the baseline in every language for both settings, exhibiting consistent results across the two settings. 
The relatively lower AUROC for Arabic stems from the poor performance of the underlying Falcon-7B scoring models in that language. Employing the stronger multilingual model davinci-002 via Glimpse \cite{bao2025glimpse} achieves significantly higher AUROC scores of 0.678 for Glimpse and 0.792 for Triospect.

\revision{
\paragraph{Across Text Domains}
We evaluated detector performance across general writing, news, essays, and scientific articles (ArXiv) under both `in-domain' and `out-of-domain' settings, where GMM parameters were estimated with or without samples from the target test domain. Table \ref{tab:results_per_domain} shows that all detectors suffered only minor drops in out-of-domain setting (average drop of 0.004 in AUROC), indicating strong robustness. Triospect consistently outperformed the baseline Fast-Detect across all domains, with notable gains except for essays, demonstrating reliable detection even on unseen text domains.
}

\revision{
\paragraph{Across Transformation Models}
We evaluated robustness across different transformation models to assess potential self-bias. When the transformation model matches the source generator, performance remains stable: using GPT-4o to transform GPT-4o–generated texts achieves nearly identical AUROC (0.814) compared to using a different transformation model, Qwen3-4B (0.810). We also swap the transformation model between calibration and testing. Calibrating the GMM with GPT-4o and testing with Qwen3-4B yields nearly identical AUROC (0.900 vs.\ 0.901) compared to using Qwen3-4B for both stages, with the same pattern observed in reverse. These results suggest that the transformation model introduces little systematic bias, and the learned densities reflect intrinsic content and expression signals rather than rewriting-model artifacts.

\subsection{Analysis of Failure Cases}
As shown in Figure \ref{fig:main_raid_analysis_barchart}, Triospect does not improve performance over the base detector under the synonym attack. We further analyze the failure cases to understand this behavior.

Specifically, Binoculars makes 108 wrong predictions, all of which are AI-generated texts after the attack, while Triospect makes 118 wrong predictions, including 99 AI-generated texts and 19 human-written texts after the attack. This result suggests that although Triospect reduces some false negatives on AI-generated texts, it also introduces additional false positives on human-written texts.

A possible reason is that synonym attacks involve only lexical substitutions while preserving the semantic content as well as stylistic and grammatical patterns. As a result, the distributions of the original text, the content-preserving text, and the expression-preserving text remain highly similar. This is evidenced by a Pearson correlation of 0.80 between $m(T)$ and $m(\hat{T}_c)$, and 0.81 between $m(T)$ and $m(\hat{T}_e)$, compared to 0.69 and 0.75 under other attacks. Such strong correlations indicate limited complementary information across the three dimensions, reducing the benefit of the multi-dimensional framework. Moreover, the rewriting transformations in Triospect may introduce additional variance, which can slightly distort human-written texts and lead to misclassification.

The key intuition is that synonym attacks modify the expression too weakly to produce meaningful differences across the three dimensions, so the additional transformations provide little benefit and may even introduce noise.

}

\section{Discussion and Limitations}
Readers might wonder how human-written texts differ from AI-generated ones once both have undergone textual transformations that make them resemble outputs from an LLM. While the way they are expressed changes, their core semantics remain the same, and this consistency can reveal their original source. \revision{Typically, human texts carry rich, specific meanings, whereas AI texts tend to be simpler and lack striking, vivid details \cite{holtzman2019curious,gehrmann2019gltr}. Differences at the semantic level will also lead to distinct and recognizable token distributions \cite{sahlgren2008distributional}}. When we preserve the semantics while surpassing other aspects, this distinct feature is displayed and can be measured using existing detection metrics.

Despite the promising results of the proposed Triospect detection framework, two limitations remain.
First, our approach relies on the approximate decoupling of content and expression through textual transformations. While effective in practice, these transformations are imperfect, leaving an open research topic regarding the disentanglement of the two aspects.
Second, the current implementation incurs additional computational costs compared to single-dimensional detectors, as multiple transformations and measurements are required. While we explored efficiency trade-offs, practical deployment in large-scale or real-time scenarios may require further optimization.
% Finally, the arms race between detectors and attackers is ongoing. As humanizing tools and adversarial methods evolve, new attack strategies may emerge that partially compromise the effectiveness of our framework. 

\section{Conclusion}
Triospect Detection Framework provides a significant advancement in AI-generated text detection, particularly against humanizing and adversarial attacks. By introducing distinct content and expression dimensions, the framework overcomes the limitations of existing detectors. Experimental results on diverse datasets demonstrate the resilience and improved performance of Triospect, achieving notable gains in AUROC and TPR01. This novel approach marks a pioneering step in enhancing the robustness and reliability of AI-generated text detection tools under attack.

\section*{Acknowledgement}
We would like to thank the editors and anonymous reviewers for their valuable feedback. This work is funded by the National Natural Science Foundation of China Key Program (Grant No. 62336006).

\bibliography{tacl2021}

@inproceedings{papineni2002bleu,
  title={Bleu: a method for automatic evaluation of machine translation},
  author={Papineni, Kishore and Roukos, Salim and Ward, Todd and Zhu, Wei-Jing},
  booktitle={Proceedings of the 40th annual meeting of the Association for Computational Linguistics},
  pages={311--318},
  year={2002}
}

@article{koppel2009computational,
  title={Computational methods in authorship attribution},
  author={Koppel, Moshe and Schler, Jonathan and Argamon, Shlomo},
  journal={Journal of the American Society for information Science and Technology},
  volume={60},
  number={1},
  pages={9--26},
  year={2009},
  publisher={Wiley Online Library}
}

@article{liu2019roberta,
  title={Roberta: A robustly optimized bert pretraining approach},
  author={Liu, Yinhan and Ott, Myle and Goyal, Naman and Du, Jingfei and Joshi, Mandar and Chen, Danqi and Levy, Omer and Lewis, Mike and Zettlemoyer, Luke and Stoyanov, Veselin},
  journal={arXiv preprint arXiv:1907.11692},
  year={2019}
}

@inproceedings{gao2021simcse,
  title={SimCSE: Simple Contrastive Learning of Sentence Embeddings},
  author={Gao, Tianyu and Yao, Xingcheng and Chen, Danqi},
  booktitle={Proceedings of the 2021 Conference on Empirical Methods in Natural Language Processing},
  pages={6894--6910},
  year={2021}
}

@article{sahlgren2008distributional,
  title={The distributional hypothesis},
  author={Sahlgren, Magnus},
  journal={Italian Journal of linguistics},
  volume={20},
  pages={33--53},
  year={2008}
}

@article{holtzman2019curious,
  title={The curious case of neural text degeneration},
  author={Holtzman, Ari and Buys, Jan and Du, Li and Forbes, Maxwell and Choi, Yejin},
  journal={arXiv preprint arXiv:1904.09751},
  year={2019}
}

@article{ilyas2019adversarial,
  title={Adversarial examples are not bugs, they are features},
  author={Ilyas, Andrew and Santurkar, Shibani and Tsipras, Dimitris and Engstrom, Logan and Tran, Brandon and Madry, Aleksander},
  journal={Advances in neural information processing systems},
  volume={32},
  year={2019}
}

@article{madry2017towards,
  title={Towards deep learning models resistant to adversarial attacks},
  author={Madry, Aleksander and Makelov, Aleksandar and Schmidt, Ludwig and Tsipras, Dimitris and Vladu, Adrian},
  journal={arXiv preprint arXiv:1706.06083},
  year={2017}
}

@inproceedings{jin2020bert,
  title={Is bert really robust? a strong baseline for natural language attack on text classification and entailment},
  author={Jin, Di and Jin, Zhijing and Zhou, Joey Tianyi and Szolovits, Peter},
  booktitle={Proceedings of the AAAI conference on artificial intelligence},
  volume={34},
  number={05},
  pages={8018--8025},
  year={2020}
}

@inproceedings{ribeiro2018semantically,
  title={Semantically equivalent adversarial rules for debugging NLP models},
  author={Ribeiro, Marco Tulio and Singh, Sameer and Guestrin, Carlos},
  booktitle={Proceedings of the 56th Annual Meeting of the Association for Computational Linguistics (volume 1: long papers)},
  pages={856--865},
  year={2018}
}

@article{alzantot2018generating,
  title={Generating natural language adversarial examples},
  author={Alzantot, Moustafa and Sharma, Yash and Elgohary, Ahmed and Ho, Bo-Jhang and Srivastava, Mani and Chang, Kai-Wei},
  journal={arXiv preprint arXiv:1804.07998},
  year={2018}
}

@article{jia2017adversarial,
  title={Adversarial examples for evaluating reading comprehension systems},
  author={Jia, Robin and Liang, Percy},
  journal={arXiv preprint arXiv:1707.07328},
  year={2017}
}

@inproceedings{zhudna2025,
  title={DNA-DetectLLM: Unveiling AI-Generated Text via a DNA-Inspired Mutation-Repair Paradigm},
  author={Zhu, Xiaowei and Ren, Yubing and Fang, Fang and Tan, Qingfeng and Wang, Shi and Cao, Yanan},
  booktitle={The Thirty-ninth Annual Conference on Neural Information Processing Systems},
  year={2025}
}

@inproceedings{maoraidar2024,
  title={Raidar: geneRative AI Detection viA Rewriting},
  author={Mao, Chengzhi and Vondrick, Carl and Wang, Hao and Yang, Junfeng},
  booktitle={The Twelfth International Conference on Learning Representations},
  year={2024}
}

@inproceedings{liu2024does,
  title={Does DetectGPT Fully Utilize Perturbation? Bridging Selective Perturbation to Fine-tuned Contrastive Learning Detector would be Better},
  author={Liu, Shengchao and Liu, Xiaoming and Wang, Yichen and Cheng, Zehua and Li, Chengzhengxu and Zhang, Zhaohan and Lan, Yu and Shen, Chao},
  booktitle={Proceedings of the 62nd Annual Meeting of the Association for Computational Linguistics (Volume 1: Long Papers)},
  pages={1874--1889},
  year={2024}
}

@article{masrour2025damage,
  title={DAMAGE: Detecting Adversarially Modified AI Generated Text},
  author={Masrour, Elyas and Emi, Bradley and Spero, Max},
  journal={arXiv preprint arXiv:2501.03437},
  year={2025}
}

@article{ayub2024art,
  title={The art of deception: humanizing AI to outsmart detection},
  author={Ayub, Taseef and Ahmad Malla, Rayees and Khan, Mashood Yousuf and Ganaie, Shabir Ahmad},
  journal={Global Knowledge, Memory and Communication},
  year={2024},
  publisher={Emerald Publishing Limited}
}

@inproceedings{zhou2024humanizing,
  title={Humanizing Machine-Generated Content: Evading AI-Text Detection through Adversarial Attack},
  author={Zhou, Ying and He, Ben and Sun, Le},
  booktitle={Proceedings of the 2024 Joint International Conference on Computational Linguistics, Language Resources and Evaluation (LREC-COLING 2024)},
  pages={8427--8437},
  year={2024}
}

@article{dwivedi2017analysis,
  title={Analysis of small sample size studies using nonparametric bootstrap test with pooled resampling method},
  author={Dwivedi, Alok Kumar and Mallawaarachchi, Indika and Alvarado, Luis A},
  journal={Statistics in medicine},
  volume={36},
  number={14},
  pages={2187--2205},
  year={2017},
  publisher={Wiley Online Library}
}

@article{mcnemar1947note,
  title={Note on the sampling error of the difference between correlated proportions or percentages},
  author={McNemar, Quinn},
  journal={Psychometrika},
  volume={12},
  number={2},
  pages={153--157},
  year={1947},
  publisher={Springer-Verlag New York}
}

@article{yu2024dpic,
  title={DPIC: Decoupling prompt and intrinsic characteristics for llm generated text detection},
  author={Yu, Xiao and Qi, Yuang and Chen, Kejiang and Chen, Guoqiang and Yang, Xi and Zhu, Pengyuan and Shang, Xiuwei and Zhang, Weiming and Yu, Nenghai},
  journal={Advances in Neural Information Processing Systems},
  volume={37},
  pages={16194--16212},
  year={2024}
}

@article{youden1950index,
  title={Index for rating diagnostic tests},
  author={Youden, William J},
  journal={Cancer},
  volume={3},
  number={1},
  pages={32--35},
  year={1950},
  publisher={Wiley Online Library}
}

@inproceedings{chen2025imitate,
  title={Imitate Before Detect: Aligning Machine Stylistic Preference for Machine-Revised Text Detection},
  author={Chen, Jiaqi and Zhu, Xiaoye and Liu, Tianyang and Chen, Ying and Xinhui, Chen and Yuan, Yiwen and Leong, Chak Tou and Li, Zuchao and Tang, Long and Zhang, Lei and others},
  booktitle={Proceedings of the AAAI Conference on Artificial Intelligence},
  volume={39},
  number={22},
  pages={23559--23567},
  year={2025}
}

@article{kintsch1978toward,
  title={Toward a model of text comprehension and production.},
  author={Kintsch, Walter and Van Dijk, Teun A},
  journal={Psychological review},
  volume={85},
  number={5},
  pages={363},
  year={1978},
  publisher={American Psychological Association}
}

@incollection{flower2016dynamics,
  title={The dynamics of composing: Making plans and juggling constraints},
  author={Flower, Linda S and Hayes, John R},
  booktitle={Cognitive processes in writing},
  pages={31--50},
  year={2016},
  publisher={Routledge}
}

@article{derose1997text,
  title={What is text, really?},
  author={DeRose, Steven J and Durand, David G and Mylonas, Elli and Renear, Allen H},
  journal={ACM SIGDOC Asterisk Journal of Computer Documentation},
  volume={21},
  number={3},
  pages={1--24},
  year={1997},
  publisher={ACM New York, NY, USA}
}

@article{clark2007content,
  title={Content management and the separation of presentation and content},
  author={Clark, Dave},
  journal={Technical communication quarterly},
  volume={17},
  number={1},
  pages={35--60},
  year={2007},
  publisher={Taylor \& Francis}
}

@article{bickhard1993representational,
  title={Representational content in humans and machines},
  author={Bickhard, Mark H},
  journal={Journal of Experimental \& Theoretical Artificial Intelligence},
  volume={5},
  number={4},
  pages={285--333},
  year={1993},
  publisher={Taylor \& Francis}
}

@article{wang2024stumbling,
  title={Stumbling blocks: Stress testing the robustness of machine-generated text detectors under attacks},
  author={Wang, Yichen and Feng, Shangbin and Hou, Abe Bohan and Pu, Xiao and Shen, Chao and Liu, Xiaoming and Tsvetkov, Yulia and He, Tianxing},
  journal={arXiv preprint arXiv:2402.11638},
  year={2024}
}

@article{zhao2024sok,
  title={SoK: Watermarking for AI-Generated Content},
  author={Zhao, Xuandong and Gunn, Sam and Christ, Miranda and Fairoze, Jaiden and Fabrega, Andres and Carlini, Nicholas and Garg, Sanjam and Hong, Sanghyun and Nasr, Milad and Tramer, Florian and others},
  journal={arXiv preprint arXiv:2411.18479},
  year={2024}
}

@article{zhao2024permute,
  title={Permute-and-Flip: An optimally robust and watermarkable decoder for LLMs},
  author={Zhao, Xuandong and Li, Lei and Wang, Yu-Xiang},
  journal={arXiv preprint arXiv:2402.05864},
  year={2024}
}

@article{bao2025glimpse,
  title={Glimpse: Enabling White-Box Methods to Use Proprietary Models for Zero-Shot LLM-Generated Text Detection},
  author={Bao, Guangsheng and Zhao, Yanbin and He, Juncai and Zhang, Yue},
  journal={The Thirteenth International Conference on Learning Representations},
  year={2025}
}

@article{guo2023hc3,
    title = "How Close is ChatGPT to Human Experts? Comparison Corpus, Evaluation, and Detection",
    author = "Guo, Biyang  and
      Zhang, Xin  and
      Wang, Ziyuan  and
      Jiang, Minqi  and
      Nie, Jinran  and
      Ding, Yuxuan  and
      Yue, Jianwei  and
      Wu, Yupeng",
    journal={arXiv preprint arxiv:2301.07597},
    year = "2023",
}

@article{dapretto1999form,
  title={Form and content: dissociating syntax and semantics in sentence comprehension},
  author={Dapretto, Mirella and Bookheimer, Susan Y},
  journal={Neuron},
  volume={24},
  number={2},
  pages={427--432},
  year={1999},
  publisher={Elsevier}
}

@article{moro2001syntax,
  title={Syntax and the brain: disentangling grammar by selective anomalies},
  author={Moro, Andrea and Tettamanti, Marco and Perani, Daniela and Donati, Caterina and Cappa, Stefano F and Fazio, Ferruccio},
  journal={Neuroimage},
  volume={13},
  number={1},
  pages={110--118},
  year={2001},
  publisher={Elsevier}
}

@inproceedings{bao2019generating,
  title={Generating Sentences from Disentangled Syntactic and Semantic Spaces},
  author={Bao, Yu and Zhou, Hao and Huang, Shujian and Li, Lei and Mou, Lili and Vechtomova, Olga and Dai, Xinyu and Chen, Jiajun},
  booktitle={Proceedings of the 57th Annual Meeting of the Association for Computational Linguistics},
  pages={6008--6019},
  year={2019}
}

@inproceedings{chen2019multi,
  title={A Multi-Task Approach for Disentangling Syntax and Semantics in Sentence Representations},
  author={Chen, Mingda and Tang, Qingming and Wiseman, Sam and Gimpel, Kevin},
  booktitle={Proceedings of the 2019 Conference of the North American Chapter of the Association for Computational Linguistics: Human Language Technologies, Volume 1 (Long and Short Papers)},
  pages={2453--2464},
  year={2019}
}

@inproceedings{caucheteux2021disentangling,
  title={Disentangling syntax and semantics in the brain with deep networks},
  author={Caucheteux, Charlotte and Gramfort, Alexandre and King, Jean-Remi},
  booktitle={International conference on machine learning},
  pages={1336--1348},
  year={2021},
  organization={PMLR}
}

@inproceedings{gao2018black,
  title={Black-box generation of adversarial text sequences to evade deep learning classifiers},
  author={Gao, Ji and Lanchantin, Jack and Soffa, Mary Lou and Qi, Yanjun},
  booktitle={2018 IEEE Security and Privacy Workshops (SPW)},
  pages={50--56},
  year={2018},
  organization={IEEE}
}

@inproceedings{dyrmishi2023humans,
  title={How do humans perceive adversarial text? A reality check on the validity and naturalness of word-based adversarial attacks},
  author={Dyrmishi, Salijona and GHAMIZI, Salah and Cordy, Maxime},
  booktitle={The 61st Annual Meeting Of The Association For Computational Linguistics},
  year={2023}
}

@inproceedings{he2024mgtbench,
  title={Mgtbench: Benchmarking machine-generated text detection},
  author={He, Xinlei and Shen, Xinyue and Chen, Zeyuan and Backes, Michael and Zhang, Yang},
  booktitle={Proceedings of the 2024 on ACM SIGSAC Conference on Computer and Communications Security},
  pages={2251--2265},
  year={2024}
}

@inproceedings{wu2024detectrl,
  title={DetectRL: Benchmarking LLM-Generated Text Detection in Real-World Scenarios},
  author={Wu, Junchao and Zhan, Runzhe and Wong, Derek F and Yang, Shu and Yang, Xinyi and Yuan, Yulin and Chao, Lidia S},
  booktitle={The Thirty-eight Conference on Neural Information Processing Systems Datasets and Benchmarks Track},
  year={2024}
}

@article{dugan2024raid,
  title={RAID: A Shared Benchmark for Robust Evaluation of Machine-Generated Text Detectors},
  author={Dugan, Liam and Hwang, Alyssa and Trhlik, Filip and Ludan, Josh Magnus and Zhu, Andrew and Xu, Hainiu and Ippolito, Daphne and Callison-Burch, Chris},
  journal={arXiv preprint arXiv:2405.07940},
  year={2024}
}

@inproceedings{hans2024spotting,
  title={Spotting LLMs With Binoculars: Zero-Shot Detection of Machine-Generated Text},
  author={Hans, Abhimanyu and Schwarzschild, Avi and Cherepanova, Valeriia and Kazemi, Hamid and Saha, Aniruddha and Goldblum, Micah and Geiping, Jonas and Goldstein, Tom},
  booktitle={Forty-first International Conference on Machine Learning},
  year={2024}
}

@inproceedings{lo-wang-2020-s2orc,
    title = "{S}2{ORC}: The Semantic Scholar Open Research Corpus",
    author = "Lo, Kyle  and Wang, Lucy Lu  and Neumann, Mark  and Kinney, Rodney  and Weld, Daniel",
    booktitle = "Proceedings of the 58th Annual Meeting of the Association for Computational Linguistics",
    month = jul,
    year = "2020",
    address = "Online",
    publisher = "Association for Computational Linguistics",
    url = "https://www.aclweb.org/anthology/2020.acl-main.447",
    doi = "10.18653/v1/2020.acl-main.447",
    pages = "4969--4983"
}

@InProceedings{Hamborg2017,
  author     = {Hamborg, Felix and Meuschke, Norman and Breitinger, Corinna and Gipp, Bela},
  title      = {news-please: A Generic News Crawler and Extractor},
  year       = {2017},
  booktitle  = {Proceedings of the 15th International Symposium of Information Science},
  location   = {Berlin},
  doi        = {10.5281/zenodo.4120316},
  pages      = {218--223},
  month      = {March}
}

@article{scott2024asap,
  title={ASAP 2.0: Automated Student Assessment Prize},
  author={Crossley, Scott and Baffour, Perpetual and King, Jules and Burleigh, Lauryn and Reade, Walter and Demkin, Maggie},
  url={https://www.kaggle.com/datasets/lburleigh/asap-2-0},
  journal={Kaggle},
  year={2024}
}

@article{arxiv2024,
  title={arXiv},
  author={arXiv},
  url={https://arxiv.org/},
  journal={Cornell University},
  year={2024}
}

@inproceedings{yang2023dna,
  title={DNA-GPT: Divergent N-Gram Analysis for Training-Free Detection of GPT-Generated Text},
  author={Yang, Xianjun and Cheng, Wei and Wu, Yue and Petzold, Linda Ruth and Wang, William Yang and Chen, Haifeng},
  year={2023},
  booktitle={The Twelfth International Conference on Learning Representations}
}

@inproceedings{verma2024ghostbuster,
  title={Ghostbuster: Detecting Text Ghostwritten by Large Language Models},
  author={Verma, Vivek and Fleisig, Eve and Tomlin, Nicholas and Klein, Dan},
  booktitle={Proceedings of the 2024 Conference of the North American Chapter of the Association for Computational Linguistics: Human Language Technologies (Volume 1: Long Papers)},
  pages={1702--1717},
  year={2024}
}

@inproceedings{li2024mage,
  title={Mage: Machine-generated text detection in the wild},
  author={Li, Yafu and Li, Qintong and Cui, Leyang and Bi, Wei and Wang, Zhilin and Wang, Longyue and Yang, Linyi and Shi, Shuming and Zhang, Yue},
  booktitle={Proceedings of the 62nd Annual Meeting of the Association for Computational Linguistics (Volume 1: Long Papers)},
  pages={36--53},
  year={2024}
}

@article{hu2023radar,
  title={Radar: Robust ai-text detection via adversarial learning},
  author={Hu, Xiaomeng and Chen, Pin-Yu and Ho, Tsung-Yi},
  journal={Advances in Neural Information Processing Systems},
  volume={36},
  pages={15077--15095},
  year={2023}
}

@article{chen2023large,
  title={Large language model in creative work: The role of collaboration modality and user expertise},
  author={Chen, Zenan and Chan, Jason},
  journal={Available at SSRN 4575598},
  year={2023}
}

@inproceedings{bagdasaryan2022spinning,
  title={Spinning language models: Risks of propaganda-as-a-service and countermeasures},
  author={Bagdasaryan, Eugene and Shmatikov, Vitaly},
  booktitle={2022 IEEE Symposium on Security and Privacy (SP)},
  pages={769--786},
  year={2022},
  organization={IEEE}
}

@article{perkins2023academic,
  title={Academic Integrity considerations of AI Large Language Models in the post-pandemic era: ChatGPT and beyond},
  author={Perkins, Mike},
  journal={Journal of University Teaching and Learning Practice},
  volume={20},
  number={2},
  year={2023}
}

@inproceedings{yuan2022wordcraft,
  title={Wordcraft: story writing with large language models},
  author={Yuan, Ann and Coenen, Andy and Reif, Emily and Ippolito, Daphne},
  booktitle={Proceedings of the 27th International Conference on Intelligent User Interfaces},
  pages={841--852},
  year={2022}
}

@article{chen2023combating,
  title={Combating misinformation in the age of llms: Opportunities and challenges},
  author={Chen, Canyu and Shu, Kai},
  journal={AI Magazine},
  year={2023},
  publisher={Wiley Online Library}
}

@inproceedings{xu2024detecting,
  title={Detecting Subtle Differences between Human and Model Languages Using Spectrum of Relative Likelihood},
  author={Xu, Yang and Wang, Yu and An, Hao and Liu, Zhichen and Li, Yongyuan},
  booktitle={Proceedings of the 2024 Conference on Empirical Methods in Natural Language Processing},
  pages={10108--10121},
  year={2024}
}

@inproceedings{zhao2023protecting,
  title={Protecting language generation models via invisible watermarking},
  author={Zhao, Xuandong and Wang, Yu-Xiang and Li, Lei},
  booktitle={International Conference on Machine Learning},
  pages={42187--42199},
  year={2023},
  organization={PMLR}
}

@inproceedings{christ2024undetectable,
  title={Undetectable watermarks for language models},
  author={Christ, Miranda and Gunn, Sam and Zamir, Or},
  booktitle={The Thirty Seventh Annual Conference on Learning Theory},
  pages={1125--1139},
  year={2024},
  organization={PMLR}
}

@article{krishna2024paraphrasing,
  title={Paraphrasing evades detectors of ai-generated text, but retrieval is an effective defense},
  author={Krishna, Kalpesh and Song, Yixiao and Karpinska, Marzena and Wieting, John and Iyyer, Mohit},
  journal={Advances in Neural Information Processing Systems},
  volume={36},
  year={2024}
}

@article{bhattacharjee2024fighting,
  title={Fighting fire with fire: can ChatGPT detect AI-generated text?},
  author={Bhattacharjee, Amrita and Liu, Huan},
  journal={ACM SIGKDD Explorations Newsletter},
  volume={25},
  number={2},
  pages={14--21},
  year={2024},
  publisher={ACM New York, NY, USA}
}

@article{sun2024trustllm,
  title={Trustllm: Trustworthiness in large language models},
  author={Sun, Lichao and Huang, Yue and Wang, Haoran and Wu, Siyuan and Zhang, Qihui and Gao, Chujie and Huang, Yixin and Lyu, Wenhan and Zhang, Yixuan and Li, Xiner and others},
  journal={arXiv preprint arXiv:2401.05561},
  year={2024}
}

@inproceedings{ippolito2020automatic,
  title={Automatic Detection of Generated Text is Easiest when Humans are Fooled},
  author={Ippolito, Daphne and Duckworth, Daniel and Callison-Burch, Chris and Eck, Douglas},
  booktitle={Proceedings of the 58th Annual Meeting of the Association for Computational Linguistics},
  pages={1808--1822},
  year={2020}
}

@article{christian2023cnet,
  title={CNET secretly used AI on articles that didn’t disclose that fact, staff say},
  author={Christian, Jon},
  journal={Futurusm, January},
  year={2023}
}

@article{m2022exploring,
  title={Exploring the role of artificial intelligence in enhancing academic performance: A case study of ChatGPT},
  author={M Alshater, Muneer},
  journal={Available at SSRN},
  year={2022}
}

@inproceedings{gehrmann2019gltr,
  title={GLTR: Statistical Detection and Visualization of Generated Text},
  author={Gehrmann, Sebastian and Strobelt, Hendrik and Rush, Alexander M},
  booktitle={Proceedings of the 57th Annual Meeting of the Association for Computational Linguistics: System Demonstrations},
  pages={111--116},
  year={2019}
}

@inproceedings{bao2024fast,
  title={Fast-DetectGPT: Efficient Zero-Shot Detection of Machine-Generated Text via Conditional Probability Curvature},
  author={Bao, Guangsheng and Zhao, Yanbin and Teng, Zhiyang and Yang, Linyi and Zhang, Yue},
  booktitle={The Twelfth International Conference on Learning Representations},
  year={2024}
}

@article{sadasivan2023can,
  title={Can AI-generated text be reliably detected?},
  author={Sadasivan, Vinu Sankar and Kumar, Aounon and Balasubramanian, Sriram and Wang, Wenxiao and Feizi, Soheil},
  journal={arXiv preprint arXiv:2303.11156},
  year={2023}
}

@article{kumar2024academic,
  title={Academic integrity and artificial intelligence: An overview},
  author={Kumar, Rahul and Eaton, Sarah Elaine and Mindzak, Michael and Morrison, Ryan},
  journal={Second handbook of academic integrity},
  pages={1583--1596},
  year={2024},
  publisher={Springer}
}

@inproceedings{lee2023language,
  title={Do language models plagiarize?},
  author={Lee, Jooyoung and Le, Thai and Chen, Jinghui and Lee, Dongwon},
  booktitle={Proceedings of the ACM Web Conference 2023},
  pages={3637--3647},
  year={2023}
}

@article{kaur2022trustworthy,
  title={Trustworthy artificial intelligence: a review},
  author={Kaur, Davinder and Uslu, Suleyman and Rittichier, Kaley J and Durresi, Arjan},
  journal={ACM Computing Surveys (CSUR)},
  volume={55},
  number={2},
  pages={1--38},
  year={2022},
  publisher={ACM New York, NY}
}

@article{ahmed2021detecting,
  title={Detecting fake news using machine learning: A systematic literature review},
  author={Ahmed, Alim Al Ayub and Aljabouh, Ayman and Donepudi, Praveen Kumar and Choi, Myung Suh},
  journal={arXiv preprint arXiv:2102.04458},
  year={2021}
}

@article{yan2023detection,
  title={Detection of AI-generated essays in writing assessment},
  author={Yan, Duanli and Fauss, Michael and Hao, Jiangang and Cui, Wenju},
  journal={Psychological Testing and Assessment Modeling},
  volume={65},
  number={2},
  pages={125--144},
  year={2023}
}

@inproceedings{pu2023deepfake,
  title={Deepfake text detection: Limitations and opportunities},
  author={Pu, Jiameng and Sarwar, Zain and Abdullah, Sifat Muhammad and Rehman, Abdullah and Kim, Yoonjin and Bhattacharya, Parantapa and Javed, Mobin and Viswanath, Bimal},
  booktitle={2023 IEEE Symposium on Security and Privacy (SP)},
  pages={1613--1630},
  year={2023},
  organization={IEEE}
}

@article{solaiman2019release,
  title={Release strategies and the social impacts of language models},
  author={Solaiman, Irene and Brundage, Miles and Clark, Jack and Askell, Amanda and Herbert-Voss, Ariel and Wu, Jeff and Radford, Alec and Krueger, Gretchen and Kim, Jong Wook and Kreps, Sarah and others},
  journal={arXiv preprint arXiv:1908.09203},
  year={2019}
}

@inproceedings{kirchenbauer2023watermark,
  title={A watermark for large language models},
  author={Kirchenbauer, John and Geiping, Jonas and Wen, Yuxin and Katz, Jonathan and Miers, Ian and Goldstein, Tom},
  booktitle={International Conference on Machine Learning},
  pages={17061--17084},
  year={2023},
  organization={PMLR}
}

@article{fagni2021tweepfake,
  title={TweepFake: About detecting deepfake tweets},
  author={Fagni, Tiziano and Falchi, Fabrizio and Gambini, Margherita and Martella, Antonio and Tesconi, Maurizio},
  journal={Plos one},
  volume={16},
  number={5},
  pages={e0251415},
  year={2021},
  publisher={Public Library of Science San Francisco, CA USA}
}

@inproceedings{uchendu2020authorship,
  title={Authorship attribution for neural text generation},
  author={Uchendu, Adaku and Le, Thai and Shu, Kai and Lee, Dongwon},
  booktitle={Proceedings of the 2020 conference on empirical methods in natural language processing (EMNLP)},
  pages={8384--8395},
  year={2020}
}

@article{bakhtin2019real,
  title={Real or fake? learning to discriminate machine from human generated text},
  author={Bakhtin, Anton and Gross, Sam and Ott, Myle and Deng, Yuntian and Ranzato, Marc'Aurelio and Szlam, Arthur},
  journal={arXiv preprint arXiv:1906.03351},
  year={2019}
}

@inproceedings{fan2018hierarchical,
  title={Hierarchical Neural Story Generation},
  author={Fan, Angela and Lewis, Mike and Dauphin, Yann},
  booktitle={Proceedings of the 56th Annual Meeting of the Association for Computational Linguistics (Volume 1: Long Papers)},
  year={2018},
  organization={Association for Computational Linguistics}
}

@inproceedings{mitchell2023detectgpt,
  title={Detectgpt: Zero-shot machine-generated text detection using probability curvature},
  author={Mitchell, Eric and Lee, Yoonho and Khazatsky, Alexander and Manning, Christopher D and Finn, Chelsea},
  booktitle={International Conference on Machine Learning},
  pages={24950--24962},
  year={2023},
  organization={PMLR}
}

@article{su2023detectllm,
  title={DetectLLM: Leveraging Log Rank Information for Zero-Shot Detection of Machine-Generated Text},
  author={Su, Jinyan and Zhuo, Terry Yue and Wang, Di and Nakov, Preslav},
  journal={arXiv preprint arXiv:2306.05540},
  year={2023}
}
\bibliographystyle{acl_natbib}

\newpage
\appendix

\section{Commercial Humanizing Tools}
\label{app:external_tools}

There are various AI humanizing tools that are developed to bypass detectors. We list a few in Table \ref{tab:list_humanizing_system}, where the first three are used to produce our humanized texts.

\begin{table}[h]
    \centering\scriptsize
    \begin{tabular}{llc}
        \toprule
        \bf AI Tool & \bf URL & \bf Used \\
        \midrule
        BypassGPT & https://bypassgpt.ai/ & Y \\
        Humbot & https://humbot.ai/ & Y \\
        Undetectable AI & https://undetectable.ai/ & Y \\
        Semihuman AI & https://semihuman.ai/ \\
        HIX Bypass & https://bypass.hix.ai/ \\
        AI Humanizer & https://aihumanizer.ai/ \\
        StealthGPT & https://stealthgpt.ai/ \\
        GPTinf & https://stealthgpt.ai/ \\
        WriteHuman & https://writehuman.ai/ \\
        StealthWriter & https://rewritify.ai/ \\
        Phrasly LLC & https://phrasly.ai/ \\
        HIX.AI & https://bypass.hix.ai \\
        AISEO Humanizer & https://aiseo.ai/ \\
        Humanize AI Pro & https://www.humanizeai.pro/ \\
        Smodin & https://smodin.io/ \\
        Rewritify & https://www.rewritify.ai \\
        \bottomrule
    \end{tabular}
    \caption{Commercial humanizing tools.}
    \label{tab:list_humanizing_system}    % \vspace{-0.2in}
\end{table}

\section{Construction of Humanize-16K}
\label{app:datasets}

We create the benchmark dataset following a strict construction process and thorough quality assurance. 

\subsection{Collection of Human-Written Texts}

\paragraph{Student Essay}  
We randomly select 1,000 essays from the Automated Student Assessment Prize (ASAP) 2.0 \cite{scott2024asap}, each accompanied by a title and a prompt. These prompts are utilized to prompt LLMs to generate corresponding essays. Additionally, metadata such as `{\it race ethnicity}', `{\it gender}', and `{\it grade level}' are recorded for potential future analyses.

\paragraph{ArXiv Intro}  
To build this dataset, we collect 1,000 computer science papers from arXiv \cite{arxiv2024} by crawling PDFs published between 2020 and 2024, randomly selecting 200 papers per year. Using S2ORC \cite{lo-wang-2020-s2orc}, the PDFs are parsed to extract titles and introductions. These titles are then used to prompt LLMs to generate new paper introductions.

\paragraph{Creative Writing}  
We randomly pull 1,000 samples from WritingPrompts \cite{fan2018hierarchical}, with each sample paired with a corresponding prompt. These prompts serve as triggers for LLMs to create new fictional stories.

\paragraph{CC News} 
For this dataset, we gather 1,000 news articles in English sourced from Common Crawl \cite{Hamborg2017}. The news headlines are used to prompt LLMs to generate full news articles.

\subsection{Produce AI-Generated Texts}
We create AI-generated texts using titles or prompts derived from human-written content. For instance, in the case of student essays, we instruct LLMs with a prompt such as: ``{\it Write a student essay (no title) in \{nwords\} words (split into \{nparagraphs\} paragraphs) based on the given title: \{title\}}''. To ensure that the generated texts closely match the average length of human-written texts, we specify the same number of words (or characters for Chinese) and paragraphs in the prompt. The detailed prompts for all domains can be found in Table \ref{tab:prompt_data_generation}.

\begin{table}[h]
    \caption{Prompts for data generation, where the {\it field} could be either `{\it title}' or `{\it prompt}' depending on their availability for each data source.}
    \label{tab:prompt_data_generation}
    \centering\small
    \begin{NiceTabular}{p{0.95\linewidth}}
        \toprule
        \textbf{Student Essay:} {\it Write a student essay (no title) in \{n\_words\} words (split into \{n\_paragraphs\} paragraphs) based on the given \{field\}:$\backslash$n \{field\_value\}} \\
        \addlinespace[0.5em]
        \textbf{ArXiv Intro:} {\it Write an introductory section (no section name) for an academic paper in \{n\_words\} words (split into \{n\_paragraphs\} paragraphs) based on the given \{field\}:$\backslash$n \{field\_value\}} \\
        \addlinespace[0.5em]
        \textbf{Creative Writing:} {\it Write a creative story (no title) in \{n\_words\} words (split into \{n\_paragraphs\} paragraphs) based on the given \{field\}:$\backslash$n \{field\_value\}} \\
        \addlinespace[0.5em]
        \textbf{CC News:} {\it Write a news article (no title) in \{n\_words\} words (split into \{n\_paragraphs\} paragraphs) based on the given \{field\}:$\backslash$n \{field\_value\}} \\
        \addlinespace[0.5em]
        \textbf{Multi-lingual CC News:} {\it Write a news article (no title)  in \{lang\} language in \{n\_words\} words (split into \{n\_paragraphs\} paragraphs) based on the given \{field\}:$\backslash$n \{field\_value\}} \\
        \bottomrule
    \end{NiceTabular}
\end{table}

We use six language models -- gpt-3.5-turbo, gpt-4o, claude-3.5-sonnet, gemini-1.5-pro, llama-3.3-70b-instruct, and qwen-2.5-72b-instruct -- to generate data, with a random model selected for each sample. In terms of decoding parameters, a temperature is randomly chosen from the range $[0.8, 1.0, 1.2]$, a top-$p$ from $[0.96, 1.0]$, and both frequency and presence penalties from the range $[0.0, 1.0]$ for each sample.

\subsection{Quality Assurance}
\label{app:quality_assure}
We evaluate the length of each generation from the LLM output, and if it is significantly longer (more than twice the original length) or shorter (less than half the original length), we prompt the LLM to generate the text again. Additionally, we monitor for issues like repetition or nonsensical responses and address them by regenerating the text. After processing the data, we truncate the texts to ensure that the length distributions are consistent across different types. As a final quality check, we randomly select 100 samples per domain for manual review, achieving an average pass rate of 99.5\%. 
% In terms of costs, the data construction process involves approximately \$2,000 for the use of the LLM API, and \$500 for manual review.

% \section{Additional Analysis}
% \label{app:analysis}

% \paragraph{Robustness to Non-Native Bias}
% Triospect measures are also resilient to nonnative English, where unique language expressions are reduced by content-preserving transformation. Consequently, using the parameters and best threshold found on the RAID development set, 2D(Binoculars) improves the AUROC from 0.497 to 0.521 and the F1 from 49\% to 55\% compared to Binoculars, demonstrating that the Triospect measures reduce the bias toward nonnative writers.

\section{Ethical Considerations}
The dataset we create contains AI-generated texts, which may occasionally exhibit bias, offensive language, or irresponsibility. While our manual review of 100 samples per domain achieved a 100\% pass rate, there remains a small possibility that some content in the broader dataset could be less than ideal. However, this risk is mitigated by the rigorous quality control measures we have implemented, and any concerns can be addressed with appropriate disclaimers or warnings.

Additionally, while AI tools are increasingly used to identify whether an article is AI-generated, we believe that these tools should be viewed as helpful aids rather than definitive authorities. Over-reliance on such technology could lead to inaccuracies or misuse, and we advocate for incorporating human judgment as an essential step in verifying the origin of any piece of work. This balanced approach ensures responsible use of AI detection tools while minimizing potential ethical concerns.

\end{document}